%% file: example_paper.tex
\documentclass{article}

\usepackage{microtype}
\usepackage{graphicx}
\usepackage{subcaption}
\usepackage{booktabs} % for professional tables

\usepackage[update]{epstopdf}

\usepackage{hyperref}

\usepackage[preprint]{icml2026}

\usepackage{amsmath}
\usepackage{amssymb}
\usepackage{mathtools}
\usepackage{amsthm}

\usepackage{acronym}
\usepackage{multirow}

\usepackage[capitalize,noabbrev]{cleveref}

\include{Wgroup-Teaching-Preamble}

\theoremstyle{plain}

\theoremstyle{definition}

\theoremstyle{remark}

\acrodef{co}[CO]{combinatorial optimization}
\acrodef{nco}[NCO]{neural combinatorial optimization}
\acrodef{rp}[RP]{routing problem}
\acrodef{rl}[RL]{reinforcement learning}
\acrodef{tsp}[TSP]{traveling salesman problem}
\acrodef{tsptw}[TSPTW]{traveling salesman problem with time window}
\acrodef{vrp}[VRP]{vehicle routing problem}
\acrodef{cvrp}[CVRP]{capacitated vehicle routing problem}
\acrodef{vrptw}[VRPTW]{vehicle routing problem with time window}
\acrodef{gnn}[GNN]{graph neural network}
\acrodef{gam}[GAM]{graph attention model}
\acrodef{gcn}[GCN]{graph convolutional network}
\acrodef{egam}[EGAM]{extended graph attention model}
\acrodef{dl}[DL]{deep learning}
\acrodef{nn}[NN]{neural network}
\acrodef{drl}[DRL]{deep reinforcement learning}
\acrodef{sota}[SOTA]{state-of-the-art}
\acrodef{or}[OR]{operations research}
\acrodef{ai4rp}[AI4RP]{Artificial Intelligence for Routing Problem}
\acrodef{pn}[PN]{Pointer Network}
\acrodef{s2v}[S2V]{Structure2Vec}
\acrodef{cnn}[CNN]{convolutional neural network}
\acrodef{mha}[MHA]{multi-head attention}
\acrodef{ml}[ML]{machine learning}
\acrodef{sl}[SL]{supervised learning}
\acrodef{ff}[FF]{feed-forward}
\acrodef{ffn}[FFN]{feed-forward network}
\acrodef{rnn}[RNN]{recurrent neural network}
\acrodef{nlp}[NLP]{natural language processing}
\acrodef{egat}[EGAT]{edge-featured GAT}
\acrodef{pip}[PIP]{Proactive Infeasibility Prevention}
\acrodef{tspdl}[TSPDL]{traveling salesman problem with draft limit}
\acrodef{pctsp}[PCTSP]{prize collecting traveling salesman problem}
\acrodef{llm}[LLM]{large language model}
\acrodef{vit}[ViT]{vision transformer}

\usepackage[textsize=tiny]{todonotes}

\icmltitlerunning{EGAM: Extended Graph Attention Model for Solving Routing Problems}

\begin{document}

\twocolumn[
\icmltitle{EGAM: Extended Graph Attention Model for Solving Routing Problems}



  \icmlsetsymbol{equal}{*}

  \begin{icmlauthorlist}
  \icmlauthor{Licheng Wang}{thu}
  \icmlauthor{Yuzi Yan}{thu}
  \icmlauthor{Mingtao Huang}{thu}
  \icmlauthor{Yuan Shen}{thu,lab}
  \end{icmlauthorlist}

  \icmlaffiliation{thu}{Department of Electronic Engineering, Tsinghua University, Beijing, China}
  \icmlaffiliation{lab}{Shanghai Artificial Intelligence Laboratory, Shanghai, China}
%   \icmlaffiliation{comp}{Company Name, Location, Country}
%   \icmlaffiliation{sch}{School of ZZZ, Institute of WWW, Location, Country}

  \icmlcorrespondingauthor{Yuan Shen}{shenyuan\_ee@tsinghua.edu.cn}
%   \icmlcorrespondingauthor{Firstname1 Lastname1}{first1.last1@xxx.edu}
%   \icmlcorrespondingauthor{Firstname2 Lastname2}{first2.last2@www.uk}

  \icmlkeywords{Combinatorial optimization, graph attention model, graph neural network, reinforcement learning, routing problem, traveling salesman problem}

  \vskip 0.3in
]



\printAffiliationsAndNotice{}  % no special notice (required even if empty)

\begin{abstract}
% \Ac{nco} solvers implemented with \acp{gnn} have introduced new approaches for solving routing problems.
\Ac{nco} solvers, implemented with \acp{gnn}, have introduced new approaches for solving routing problems.
Trained with \ac{rl}, the state-of-the-art \ac{gam} achieves near-optimal solutions without requiring expert knowledge or labeled data.
% In this work, we extend the attention mechanism on graph-structured data and propose the \ac{egam}.
In this work, we generalize the existing graph attention mechanism
and propose the \ac{egam}.
Our model utilizes multi-head dot-product attention to update both node and edge embeddings, addressing the limitations of the conventional \ac{gam}, which considers only node features.
We employ an autoregressive encoder–decoder architecture and train it with policy gradient algorithms that incorporate a specially designed baseline.
Experiments show that \ac{egam} matches or outperforms existing methods across various routing problems. % or 持平
% Additionally, the proposed model shows particularly strong performance on highly constrained problems, with a 2\% improvement on TSPTW and a 3\% improvement on VRPTW. % TSPDL
% The proposed model shows particularly strong performance on highly constrained problems, indicating its efficiency to handle complex graph structures.
Notably, the proposed model demonstrates exceptional performance on highly constrained problems, highlighting its efficiency in handling complex graph structures.
\end{abstract}

\acresetall

\section{Introduction}
Routing problems, such as the well-known \ac{tsp}, represent a fundamental class of \ac{co} problems.
Due to the NP-hard nature of routing problems, exact algorithms become computationally prohibitive for large-scale instances \cite{CapCheKhaLodMorVel:J23,AlaMenMoh:J25}. 
While hand-crafted heuristics can provide near-optimal solutions for specific problems, they require significant manual effort and lack generalizability \cite{KimParKim:C21,BiMaZho:C24}. 
Recently, \ac{nco} solvers, implemented with \acp{gnn}, have gained increasing attention due to their ability to autonomously generate effective heuristics \cite{KoovanWel:C19,WuWanWen:A24}. 
% 优点，经过训练，兼具推理速度和性能
Through \ac{sl} or \ac{rl}, well-trained \ac{nco} solvers strike a strong balance between solution quality and computational efficiency \cite{KwoChoKimYooGwoMin:C20,SunYan:C23}.

From a structural perspective, \ac{nco} solvers are generally classified into two categories:  autoregressive and non-autoregressive \cite{WuWanWen:A24}. 
Autoregressive solvers generate solutions sequentially and are typically trained by \ac{rl} \cite{BiMaZho:C24}. 
% In contrast, non-autoregressive solvers output a heatmap of edge-selection preferences, which requires a search procedure, e.g., Monte Carlo Tree Search \cite{Cou:J06}, to form a route \cite{SunYan:C23}. 
In contrast, non-autoregressive solvers output a heatmap of edge-selection preferences, which requires a subsequent search procedure\textemdash{}such as Monte Carlo Tree Search \cite{Cou:J06}\textemdash{}to form the final route \cite{SunYan:C23}.
Non-autoregressive solvers are usually trained using \ac{sl} \cite{JosCapRouLau:J22}.
However, \ac{sl} is only applicable to problems with labeled data, which is not universally available for all routing scenarios.
% can only be applied to problems with labeled data, which is not universally available for all routing problems. 
% Given the complex constraints and dynamic nature of real-world route planning, we argue that \ac{rl}-based autoregressive solvers are better suited to practical applications.
Given the complex constraints and dynamic nature of real-world route planning, \ac{rl}-based autoregressive solvers are better suited for practical applications due to their portability in training and deployment \cite{KwoChoKimYooGwoMin:C20,KimParPar:C22}.

Among the \ac{rl}-based \ac{nco} solvers, the \ac{gam} \cite{DeuCouLacAduRou:J18,KoovanWel:C19} is considered as the state-of-the-art architecture for routing problems. 
Inspired by Transformer \cite{VasShaParUszJonGomKaiPol:C17}, \acp{gam} utilize \ac{mha} to aggregate node features over graphs and generate solutions \cite{KoovanWel:C19}. 
% 加一个更多的研究也围绕这个结构展开
% The follow-up works POMO \cite{KwoChoKimYooGwoMin:C20} and Sym-NCO \cite{KimParPar:C22}
Subsequent works, such as POMO \cite{KwoChoKimYooGwoMin:C20} and Sym-NCO \cite{KimParPar:C22}, have further enhanced the performance of \acp{gam} by innovating its training algorithm. 
The Transformer architecture is now widely applied across various machine learning areas, including \acp{llm} in \ac{nlp} \cite{NavKhaQiuSaqAnwUsmAkhBarMia:J25} and \acp{vit} in computer vision \cite{HanWanChe:J22}. 
This highlights the effectiveness of the attention mechanisms (particularly multi-head dot-product attention) in extracting representative features from data. 
However, existing \acp{gam} mainly focus on self-attention between node representations, overlooking the crucial role of edge attributes. 
This is a significant limitation for problems where edge features are integral, such as the \ac{tsp} with edge costs. 
To address this, we design a novel architecture, the \ac{egam}, where both node and edge embeddings are updated via attention mechanism. 

% So what is attention? 
% From a human perspective, attention means getting information selectively from what we have seen \cite{HanWanChe:J22}. 
From a biological perspective, attention is the process of selectively acquiring information from the environment \cite{HanWanChe:J22}.
% 有侧重的获取信息？
% Given that attention operations can be applied between nodes in a graph, why not extend this mechanism to operate between nodes and edges? 
Since traditional graph attention methods are limited to operations between nodes \cite{KimParPar:C22,BiMaZho:C24}, why not extend this mechanism to operate between nodes and edges?
% In our model, we define the Node-Edge and Edge-Node Attention operations, which is for the first time, to our knowledge, that node and edge features are incorporated by multi-head dot-product attention. 太绝对了，不要这么说
% In our model, we introduce the Node-Edge and Edge-Node Attention operations, enabling bidirectional information flow between nodes and edges through multi-head dot-product attention. 
In our model, we introduce the Node-Edge and Edge-Node Attention operations, where node and edge features are fused through multi-head dot-product attention. 
Building on these attention mechanisms, we design a model architecture composed of a graph encoder and an autoregressive decoder. 
The proposed \ac{egam} 
% can perform end-to-end processing from problem input to route generation. 
is capable of performing end-to-end processing, from problem input to route generation.
% For training, we employ a baseline generation method that incorporates graph symmetries, and train the model parameters using a policy gradient algorithm.
For training, we use the REINFORCE algorithm \cite{Wil:J92} with a symmetry-based baseline for gradient estimation, enabling parameter training without the need for ground truth.

Compared to traditional \acp{gam}, the \ac{egam} demonstrates advantages in both performance and scalability.
We conducted experiments on various routing problems, categorizing them into two types based on the complexity of their constraints.
% The proposed model extends the range of problems that traditional GAMs can address. 
% Moreover, experiments demonstrate that our model achieves particularly significant performance improvements on complex routing problems. 
% Furthermore, \ac{egam} demonstrates a clear performance advantage over other \ac{rl}-based \ac{nco} solvers. 
% We conducted experiments on various routing problems. 
The results show that for problems with simple constraints, \ac{egam} performs similarly to, or slightly better than, state-of-the-art methods. 
For highly constrained problems, \ac{egam} significantly outperforms state-of-the-art approaches in terms of both solution cost and feasible rate. 
In the experiments on \ac{tsptw}, \ac{tspdl}, and \ac{vrptw}, our model reduced the optimality gap by 2.29\%, 2.21\%, and 2.04\%, respectively, under single-decision settings.
% 要不要在这里说TW后面再想
% Contributions:
% This demonstrates \ac{egam}'s efficiency in handling complex graph structures. 
These results highlight the efficiency of \ac{egam} in handling complex graph structures.
% Additionally, the introduction of edge features in the network structure expands the range of solvable problems and helps model complex transition relationships.
% In summary, we propose a novel neural network framework for routing problems. 
% The efficiency and scalability of the proposed model hold promise for offering new solutions to a broader range of NP-hard \ac{co} problems.
Furthermore, the integration of edge features into the network architecture not only expands the range of solvable problems but also enhances the modeling of complex transition relationships.
In summary, we introduce a novel neural network framework for routing problems. 
The proposed model’s efficiency and scalability provide a promising path toward addressing a broader range of NP-hard \ac{co} problems.

\section{Related Work}
\subsection{Neural combinatorial optimization}
The application of \ac{nco} dates back to Hopfield-networks \cite{HopTan:J85} and deformable template models \cite{For:J88,AngVauLe:J88}. 
Recent advancements in neural models and training methods have made NCO a competitive approach for solving CO, particularly in routing problems \cite{GarCebMen:J24}.
The Pointer Network \cite{VinForJai:C15} used a \ac{rnn} with an attention-based output layer to direct the model's focus to specific positions in the input sequence.
Building on this, \citeauthor{BelPhaLeNorBen:A16} \yrcite{BelPhaLeNorBen:A16} introduced an Actor-Critic algorithm that trained Pointer Networks without labeled data.
Additionally, Structure2Vec \cite{DaiDaiSon:C16, KhaDaiZhaDilSon:C17} employed a neural network to encode graph nodes into high-dimensional embedding vectors, which was trained by Q-learning. 
% In the development of \ac{nco} methods, neural network architectures have played an important role.
% However, these early methods fall short of traditional heuristics on both speed and solution quality \cite{la2012comparison}.
% Research over the past decade has significantly advanced neural models and training methods, enabling \ac{nco} to be a competitive approach for solving \ac{co} especially routing problems \cite{GarCebMen:J24}.
% , which pointed the output to a specific position in the input sequence. 
% further introduced an Actor-critic algorithm that trained Pointer Networks without labeled data.
% Drawing inspiration from the Transformer architecture \cite{VasShaParUszJonGomKaiPol:C17} in \ac{nlp}, \citeauthor{DeuCouLacAduRou:J18} \yrcite{DeuCouLacAduRou:J18} and \citeauthor{KoovanWel:C19} \yrcite{KoovanWel:C19} first apply \acp{gam} to routing problems. \Acp{gam} \cite{DeuCouLacAduRou:J18,KoovanWel:C19} employ a \ac{mha}-based encoder to embed graph nodes and an autoregressive decoder to output solutions.

Drawing inspiration from the Transformer architecture \cite{VasShaParUszJonGomKaiPol:C17}, \citeauthor{DeuCouLacAduRou:J18} \yrcite{DeuCouLacAduRou:J18} first applied \acp{gam} to routing problems, which consisted of a \ac{mha}-based encoder to embed graph nodes and an autoregressive decoder to generate solutions. 
\citeauthor{KoovanWel:C19} \yrcite{KoovanWel:C19} enhanced this framework by refining the decoder and incorporating a policy gradient algorithm that used a rollout baseline.
Subsequent research focused on improving the training algorithms of GAMs.
\citeauthor{KwoChoKimYooGwoMin:C20} \yrcite{KwoChoKimYooGwoMin:C20} introduced POMO, which utilizes multiple decision branches to generate a shared baseline, while \citeauthor{KimParPar:C22} \yrcite{KimParPar:C22} proposed Sym-NCO, a regularizer-based training scheme that leverages inherent symmetry in routing problems.
% These advancements have enabled \acp{gam} to achieve near-optimal solutions for routing problems without the need for labeled data, making them state-of-the-art RL-based NCO solvers \cite{CapCheKhaLodMorVel:J23}.
These advancements have further improved the performance of \acp{gam} in solving routing problems, making them state-of-the-art \ac{rl}-based NCO solvers \cite{CapCheKhaLodMorVel:J23}.
% \citeauthor{KwoChoKimYooGwoMin:C20} \yrcite{KwoChoKimYooGwoMin:C20} introduce the POMO method that averages solutions originating from multiple decision-starting branches. 
% POMO \cite{KwoChoKimYooGwoMin:C20} introduced a novel baseline that averages solutions originating from multiple decision-starting branches. 
% Additionally, \citeauthor{KimParPar:C22} \yrcite{KimParPar:C22} proposed Sym-NCO, a regularizer-based training scheme that leverages universal symmetricities inherent in routing problems. 
% Equipped with POMO \cite{KwoChoKimYooGwoMin:C20} or Sym-NCO \cite{KimParPar:C22}, GAMs have reduced the optimality gap on TSP100 to 1\% under a single trajectory. 
% Equipped with POMO \cite{KwoChoKimYooGwoMin:C20} or Sym-NCO \cite{KimParPar:C22}, GAMs achieve near-optimal solutions without requiring labeled data.
% Notably, their performance can be further enhanced through sampling. 
% Therefore, \Acp{gam} are recognized as the state-of-the-art \ac{rl}-based \ac{nco} solvers for routing problems \cite{CapCheKhaLodMorVel:J23}.

In contrast to autoregressive approaches, non-autoregressive architectures do not directly generate solutions but instead produce an adjacency matrix of edge selection preferences \cite{CapCheKhaLodMorVel:J23}.
\citeauthor{JosLauBre:A19} \yrcite{JosLauBre:A19} used \acp{gcn} to output adjacency matrices, with solutions obtained through beam search. 
Similarly, \citeauthor{KoovanGroWel:J22} \yrcite{KoovanGroWel:J22} introduced a dynamic programming-based decoding strategy, validated on \ac{tsp} and its variants. 
Adopting a graph-based diffusion model \cite{HoJaiAbb:c20}, \citeauthor{SunYan:C23} \yrcite{SunYan:C23} proposed DIFUSCO, which achieves near-optimal performance on small-scale TSP instances and maintains its effectiveness on large-scale problems. 
However, the training of non-autoregressive models relies on labeled data, which is difficult to obtain for complex routing problems.
% To enhance the efficiency of supervised training, \citeauthor{SunYan:C23} \yrcite{SunYan:C23} presented DIFUSCO, a diffusion model \cite{HoJaiAbb:c20} where the denoising network was composed of an anisotropic \ac{gcn} \cite{JosCapRouLau:J22}. 
% DIFUSCO achieves performance that closely matches that of the exact algorithm on small-scale TSP instances, while notably maintaining its effectiveness on large-scale problems.
% Nevertheless, the training of non-autoregressive models mainly depends on labeled data, which is difficult to acquire for complex routing problems.

Additionally, some studies have focused on enhancing the scalability of existing \ac{nco} solvers across diverse problem settings and scales.
\citeauthor{BiMaZho:C24} \yrcite{BiMaZho:C24} introduced the \ac{pip} framework, which enhances decision-making by incorporating a learnable preventative infeasibility mask, showing strong performance on constrained problems like \ac{tsptw} and \ac{tspdl}. 
To improve generalization for large-scale instances, \citeauthor{FuQiuZha:C21} \yrcite{FuQiuZha:C21} proposed a graph sampling-based method to augment pre-trained models. 
Moreover, \citeauthor{GaoShaXue:C25} \yrcite{GaoShaXue:C25} introduced a general selector to allocate problem instances to the most suitable models.
% For problems with complex constraints, \citeauthor{BiMaZho:C24} \yrcite{BiMaZho:C24} introduced the \ac{pip} framework. \Ac{pip} enhances the decision-making stage of \ac{nco} by incorporating a learnable preventative infeasibility mask, which has demonstrated strong performance on problems with strict visit-order requirements such as the \ac{tsptw} and the \ac{tspdl}. 
% To ensure generalization capability on large-scale instances, \citeauthor{FuQiuZha:C21} \yrcite{FuQiuZha:C21} proposed a graph sampling-based method to augment small pre-trained models.
% Moreover, to better coordinate multiple NCO solvers, \citeauthor{GaoShaXue:C25} \yrcite{GaoShaXue:C25} proposed a general selector designed to allocate different problem instances to their most suitable models.

\subsection{Attention mechanisms}
% 这个小节需要讲点公式，把注意力机制的定义，方式方法讲清楚，方便后续展开
Attention mechanisms were initially introduced in computer vision as an imitation of the human visual system \cite{mnih2014recurrent,GuoXuLiuLiuJiaMuZhaMarCheHu:J22}.
In neural machine translation, \citeauthor{BahChoBen:C15} \yrcite{BahChoBen:C15} proposed an attention-based \ac{rnn} framework, where the decoder employed an additive attention mechanism to selectively retrieve information from the embedded sequence.
Base on this framework, \citeauthor{LuoPhaMan:C15} \yrcite{LuoPhaMan:C15} generalized the mathematical operation of attention mechanisms, introducing the widely used dot-product attention \cite{NiuZhoYu:J21}. 
Aside from its application in the decoder, attention mechanisms have also been used in \acp{rnn} to capture the relationships between input tokens \cite{CheDonLap:C16,ParTacDasUsz:J16}.
Different from previous models, Transformer \cite{VasShaParUszJonGomKaiPol:C17} presented an architecture eschewing recurrence and instead relying on attention operations for sequence-to-sequence processing. 
Building on dot-product attention, \citeauthor{VasShaParUszJonGomKaiPol:C17} \yrcite{VasShaParUszJonGomKaiPol:C17} introduced \ac{mha}, which was considered superior to single-head attention for extracting multi-dimensional features. 
With the input features $\RM{X}=[\RV{x}_1,\RV{x}_2,\cdots,\RV{x}_M]$, $\RM{Y}=[\RV{y}_1,\RV{y}_2,\cdots,\RV{y}_N]$, and $\RM{Z}=[\RV{z}_1,\RV{z}_2,\cdots,\RV{z}_N]$, the  \ac{mha} is defined by the following formulas:
% 是不是应该用大H好一些？
{\setlength{\abovedisplayskip}{5pt}
\setlength{\belowdisplayskip}{5pt}
\begin{align}
\operatorname{MHA}(\RM{X}, \RM{Y},\RM{Z})=\M{W}^{\rm O}\operatorname{Concat}(\RM{H}_1,\RM{H}_2,\ldots,\RM{H}_h),\\
\RM{H}_i = \operatorname{Attention}(\RM{X}^{\rm T}\M{W}^{\rm Q}_i,\RM{Y}^{\rm T}\M{W}^{\rm K}_i,\RM{Z}^{\rm T}\M{W}^{\rm V}_i)^{\rm T}.
\end{align}}%
where the attention operation is given by
\begin{equation}\label{attention}
\setlength{\abovedisplayskip}{5pt}
\setlength{\belowdisplayskip}{5pt}
\operatorname{Attention}(\RM{Q},\RM{K},\RM{V})=\operatorname{softmax}\left(\frac{\RM{Q} \RM{K}^{\rm T}}{\sqrt{d_{\rm k}}}\right) \RM{V}.
\end{equation}
In \eqref{attention}, $d_{\rm k}$ denotes the last dimension of matrix $\RM{K}$. 
Given that the key and value inputs are often identical, i.e. $\RM{Y}=\RM{Z}$, we use $\operatorname{MHA}(\RM{X}, \RM{Y})$ to represent this case. 
% in which the additive attention mechanism was employed by the decoder to selectively retrieve information from the embedded sequence. 
% Beyond its application at the output layer, the attention mechanism also served to capture the relationships between input tokens \cite{CheDonLap:C16,ParTacDasUsz:J16}.
% In these studies, attention mechanisms commonly act as a component of \acp{rnn}.
% The Transformer architecture has achieved remarkable success in various fields of machine learning. ...

The Transformer architecture has achieved remarkable success across multiple machine learning domains.
Transformer-based models, such as GPT \cite{RadNarSal:B18} and BERT \cite{DevChaLee:C19}, have set new benchmarks in \ac{nlp} \cite{NavKhaQiuSaqAnwUsmAkhBarMia:J25}.
Models like \ac{vit} \cite{AleLucAle:C21} and DETR \cite{CarMasSym:J20} have demonstrated superior performance in computer vision tasks, including image classification and object detection \cite{HanWanChe:J22}.
The architecture's scalability and ability to capture long-range dependencies have also driven advancements in time-series forecasting, embodied learning, and protein structure prediction \cite{WanWuDon:C24,KhaNooAwa:J25,SfeHuaLiu:C25}.

% The Transformer architecture has achieved remarkable success in various fields of machine learning. 
% Transformer-based models have revolutionized language modeling, with large-scale models like GPT \cite{RadNarSal:B18} and BERT \cite{DevChaLee:C19} setting new benchmarks in tasks such as machine translation, text generation, and question answering \cite{NavKhaQiuSaqAnwUsmAkhBarMia:J25}.
% In addition to \ac{nlp}, Transformer architectures have been increasingly applied in computer vision, where models like \ac{vit} \cite{AleLucAle:C21} and DETR \cite{CarMasSym:J20} have demonstrated superior performance in image classification and object detection \cite{HanWanChe:J22}. 
% The architecture's scalability and ability to model long-range dependencies have also made it a powerful tool in time-series forecasting, embodied learning, and protein structure prediction, expanding its impact across diverse fields \cite{WanWuDon:C24,KhaNooAwa:J25,SfeHuaLiu:C25}.
% Furthermore, Transformers have shown promise in multimodal processing, where they effectively combine information from text, images, and other modalities (to be cited). 

\subsection{Graph attention networks}
The concept of graph attention network was first proposed by \citeauthor{VelCucCasRomLioBen:C18} \yrcite{VelCucCasRomLioBen:C18} for graph node classification. In their model, named GATs, the additive attention mechanism served as the primary computational tool for processing graph-structured data \cite{VelCucCasRomLioBen:C18}.
Later, \citeauthor{BroAloYah:C22} \yrcite{BroAloYah:C22} modified the additive attention mechanism and introduced GATv2. 
\acp{gam} are considered as graph attention networks tailored for routing problems \cite{KoovanWel:C19}. 
Differently, \acp{gam} utilize multi-head dot-product attention to embed graph nodes and generate routing sequences \cite{DeuCouLacAduRou:J18,KoovanWel:C19}.
Other Transformer-based architectures for routing tasks \cite{KaeWol:A18,BreLau:A21} are also categorized alongside GAMs.
% additive attention functioned as the primary computational mechanism for processing graph-structured data \cite{VelCucCasRomLioBen:C18}. 
% Differently, in \acp{gam}, the multi-head dot-product attention is employed  to embed graph nodes and output the route sequence \cite{DeuCouLacAduRou:J18,KoovanWel:C19}. 
% Additionally, other Transformer-based architectures \cite{KaeWol:A18,BreLau:A21} developed for routing problems can be roughly grouped under the same category as GAMs.

While traditional GATs focus on node representations, \citeauthor{WanCheChe:J21} \yrcite{WanCheChe:J21} proposed the \ac{egat}, which integrates edge features into the message-passing process by updating edge embeddings using a node-transit strategy.
Additionally, \citeauthor{LisRydWu:A24} \yrcite{LisRydWu:A24} introduced GREAT, an edge-focused model that incorporates edge information without relying on node embeddings.
In contrast, our work extends existing node-based GAMs in a more flexible manner. The proposed \ac{egam} relies on multi-head dot-product attention and defines multiple attention layers for different types of embeddings.
% In \ac{egat}, edge embeddings are updated via a node-transit strategy, which applies the attention mechanism between nodes.
% integrates edge information without relying on node embeddings.
% We will introduce the model architecture and training algorithm in the next section.
% The traditional graph attention networks mainly focus on node representations. \citeauthor{WanCheChe:J21} \yrcite{WanCheChe:J21} proposed the \ac{egat}, where edge features are considered in the message passing over graphs.
% \ac{egat} exhibits notable performance on node-sensitive and edge-sensitive datasets \cite{WanCheChe:J21}.
% Similarly, our work introduces new attention operations for edge features to extend the existing \ac{gam}. 
% Different from \ac{egat}, we adopt multi-head dot-product attention as the primary computation module in both the encoder and the decoder. 
% In contrast, we directly employ attention mechanisms for information fusion by defining the Edge-Node Attention and the Node-Edge Attention.
% 这一节必须从前两小节中把话题拉回来，紧扣routing problem和am，但是也不要讲太多细节
% EGAT and our model
% 需要补充最新看到的那篇文章了GREAT

\section{Methodology}
\subsection{Generalized attention mechanism over graphs}
% 可以插一张图；没空，不插了
We begin by introducing new attention operations for graph-structured data. 
Conventional \acp{gam} primarily focus on node features. However, edge features also constitute a fundamental aspect of graphs.
For example, in the \ac{tsp} with edge cost, the travelling cost is defined associated with edges, not simply the the distances between nodes.
Furthermore, we argue that incorporating edge features can also contribute to solving complex routing problems, such as those with time window constraints.
Therefore, we define two novel attention operations: Edge-Node Attention and Node-Edge Attention, which facilitate bidirectional message passing between nodes and edges.

For a full-connected graph $G=({\Set V},{\Set E})$ consisted of $N$ nodes, we assume ${\Set V}=\{n_i\mid i=1,2,\ldots,N\}$ and ${\Set E}=\{e_{i,j}\mid i,j=1,2,\ldots,N\}$. Let $\RV{n}_i^\ell$ and $\RV{e}_{i,j}^\ell$ represent the embeddings of node $n_i$ and edge $e_{ij}$, respectively, as the output of layer $\ell$.

\textbf{Node-Node Attention} is the primary operation in existing \acp{gam} for learning node embeddings \cite{DeuCouLacAduRou:J18,KoovanWel:C19}. The node embeddings are updated as follows:
\begin{equation}
\setlength{\abovedisplayskip}{5pt}
\setlength{\belowdisplayskip}{5pt}
\RV{n}_i^{\ell}=\operatorname{MHA}(\RV{n}_i^{\ell-1}, [\RV{n}_1^{\ell-1},\RV{n}_2^{\ell-1},\cdots,\RV{n}_N^{\ell-1}]).
\label{nna}
\end{equation}
For simplicity, we omit the residual connections \cite{HeZhaRen:C16} and normalization layers \cite{IofSze:C15,UlyVedLem:A16}. \ac{mha} operation is permutation-invariant to its input, meaning it is independent of node ordering, making it well-suited for graph-structured data.
% Following the \ac{mha}, a \ac{ffn} is usually applied as a non-linear transformation to each element, enhancing the model's representational capacity.

\textbf{Node-Edge Attention} operates between a node and its adjacent edges.
% As shown in \eqref{nea}, % 从节点到边的选择
% 有向图和无向图
% We embed both nodes and edges into high-dimensional space $\mathbb{R}^{d_{\rm m}}$, thus enabling the \ac{mha} operation as shown in \eqref{nea}.
% Through Node-Edge Attention, node embeddings can be updated selectively according to their adjacent edges. 
% Specifically, for undirected graphs, we assume that $\RV{e}_{i,j}^\ell=\RV{e}_{j,i}^\ell$.
Both nodes and edges are embedded into a high-dimensional space $\mathbb{R}^{d_{\rm m}}$, enabling the use of the \ac{mha} operation as shown in \eqref{nea}.
Through Node-Edge Attention, nodes selectively acquire information from their adjacent edges.
For undirected graphs, we assume symmetry in the edge embeddings, i.e., $\RV{e}_{i,j}^\ell=\RV{e}_{j,i}^\ell$.
Additionally, self-loops (e.g., $\RV{e}_{i,i}^\ell$) are incorporated to allow nodes to attend to their own information, although such edges are explicitly prohibited from being selected during decoding.
\begin{equation}
\setlength{\abovedisplayskip}{5pt}
\setlength{\belowdisplayskip}{5pt}
\RV{n}_{i}^{\ell}=\operatorname{MHA}(\RV{n}_{i}^{\ell-1}, [\RV{e}_{i,1}^{\ell-1},\RV{e}_{i,2}^{\ell-1},\cdots,\RV{e}_{i,N}^{\ell-1}]).
\label{nea}
\end{equation}

\textbf{Edge-Node Attention} is the complementary operation to Node-Edge Attention.
In undirected graphs, Edge-Node Attention is defined as:
\begin{equation}
\setlength{\abovedisplayskip}{5pt}
\setlength{\belowdisplayskip}{5pt}
\RV{e}_{i,j}^{\ell}=\operatorname{MHA}(\RV{e}_{ij}^{\ell-1}, [\RV{n}_i^{\ell-1},\RV{n}_j^{\ell-1}]).
\label{ena1}
\end{equation}
% 起终点encoding
For directed graphs, additional encoding is required to distinguish between the start and end nodes. Without loss of generality, we use separate linear layers to achieve this distinction:
\begin{equation}
\setlength{\abovedisplayskip}{5pt}
\setlength{\belowdisplayskip}{5pt}
\RV{e}_{i,j}^{\ell}=\operatorname{MHA}(\RV{e}_{ij}^{\ell-1}, [\M{W}_{\rm s}\RV{n}_i^{\ell-1}+\M{b}_{\rm s},\M{W}_{\rm e}\RV{n}_j^{\ell-1}+\M{b}_{\rm e}]).
\label{ena2}
\end{equation}

% 非全连接
% Moreover, the proposed attention mechanism is also applicable to non-fully-connected graphs by adding masks to \ac{mha}. By defining the above operations, we achieve information propagation and representation of graph elements solely through attention operations. In contrast to traditional linear and concatenation operations
% % 这地方得解释一下，提一下GCN等模型
% , we argue that the proposed graph attention mechanism enables more efficient representation of graph structures.
Node-Edge Attention and Edge-Node Attention enable bidirectional information propagation between nodes and edges through \ac{mha}. 
In contrast to the linear and activation operations in \acp{gcn} \cite{JosCapRouLau:J22}, we argue that \ac{mha} more efficiently performs the fusion and interaction of multi-dimensional features.
The proposed graph attention mechanism facilitates a more effective representation of graph structures while also being used during the decoding phase to capture dynamic context information. 
Additionally, this attention mechanism can be easily adapted to non-fully-connected graphs by incorporating masks into the \ac{mha}.

\subsection{Extended graph attention model}

\begin{figure*}[!ht]
\begin{minipage}[c]{1.0\linewidth}
  \centering
  \centerline{\includegraphics[width=0.9\textwidth]{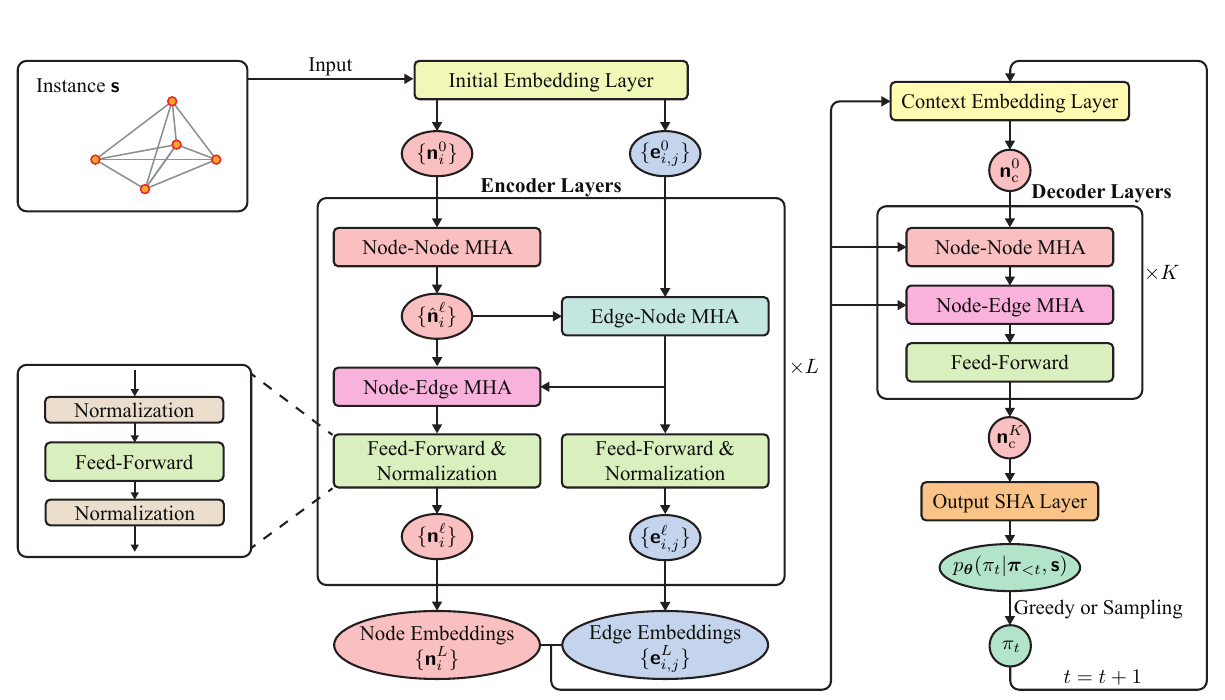}}
\end{minipage}
\caption{Illustration of the proposed \ac{egam} structure. The rounded rectangles represent trainable network layers, while the ellipses denote the output variables. For simplicity, the residual connections of the MHA and feed-forward layers are omitted. Arrows indicate the data flow from the problem input to the sequence generation. In a single decision process, the encoder runs only once, while the decoder operates repeatedly until the full route is generated.}%省略了skip-connection
\label{fig:model}
\end{figure*}

% 结合路由问题，描绘出模型的细节（含计算公式和具体的维度）
Based on the aforementioned attention mechanism, we design a new autoregressive architecture for routing problems.
The proposed \ac{egam} adopts an encoder-decoder framework, where the encoder embeds the input graph, and the decoder sequentially outputs the node sequence.

% RP
For a routing problem instance, denoted as $\RV{s}$, we represent its topological structure using a graph $G=({\Set V},{\Set E})$. The initial node features and edge features are respectively denoted as $\RV{n}_i^{\rm init}\in\mathbb{R}^{F_{\rm n}}$ and $\RV{e}_{i,j}^{\rm init}\in\mathbb{R}^{F_{\rm e}}$, respectively.
The goal is to find a route (solution) $\V{\pi}$ that minimizes the cost $C(\V{\pi}, \RV{s})$, where $\V{\pi}=[\pi_1,\pi_2,\cdots, \pi_T], \pi_t\in {\Set V}, t=1,2,\ldots, T$. 
In autoregressive \ac{nco} solvers \cite{KoovanWel:C19, BiMaZho:C24}, the output of the neural network is modeled as a conditional probability distribution $p_{\V{\theta}}(\pi_t|\V{\pi}_{<t},\RV{s})$, where $\V{\theta}$ denotes the network parameters. 

The structure of \ac{egam} is illustrated in \cref{fig:model}. 
% The encoder of \ac{egam} embeds the nodes and edges of the input graph into a high-dimensional space. 
The role of the encoder in \ac{egam} is to map the graph elements into a high-dimensional space, providing embeddings for the decoder's decision-making process.
The input features are first transformed by a linear embedding layer:
% {\setlength{\abovedisplayskip}{5pt}
% \setlength{\belowdisplayskip}{5pt}
% \begin{align}
% \RV{e}_{i,j}^{0} &= \M{W}_{\rm init}^{\rm e}\RV{e}_{i,j}^{\rm init}+\M{b}_{\rm init}^{\rm e},\\
% \RV{n}_{i}^{0} &= \M{W}_{\rm init}^{\rm n}\RV{n}_{i}^{\rm init}+\M{b}_{\rm init}^{\rm n}.
% \end{align}}%
{\setlength{\abovedisplayskip}{5pt}
\setlength{\belowdisplayskip}{5pt}
\begin{align}
\RV{n}_{i}^{0} = \M{W}_{\rm n}^{\rm init}\RV{n}_{i}^{\rm init}+\M{b}_{\rm n}^{\rm init},\\
\RV{e}_{i,j}^{0} = \M{W}_{\rm e}^{\rm init}\RV{e}_{i,j}^{\rm init}+\M{b}_{\rm e}^{\rm init},
\end{align}}%
where $\M{W}_{\rm n}^{\rm init}\in\mathbb{R}^{d_{\rm m}\times F_{\rm n}},\M{W}_{\rm e}^{\rm init}\in\mathbb{R}^{d_{\rm m}\times F_{\rm e}}, \M{b}_{\rm n}^{\rm init},\M{b}_{\rm e}^{\rm init}\in\mathbb{R}^{d_{\rm m}}$.
We set the embedding dimension $d_{\rm m}=128$. After the initial layer, the node and edge embeddings are updated using the graph attention mechanism formalized in \cref{nna,nea,ena1,ena2}.
% Our model utilizes a integrated attention layer that sequentially processes by sequentially processing Node-Node, Edge-Node, and Node-Edge attention operations. 
Our model employs integrated attention layers that sequentially processes Node-Node, Edge-Node, and Node-Edge attention operations.
We retain the existing Node-Node attention to ensure efficient information propagation across nodes.
% The newly introduced Edge-Node and Node-Edge attention mechanisms incorporate edge embeddings into the process of graph representation learning.
The newly introduced Edge-Node and Node-Edge attention mechanisms incorporate edge embeddings into the graph representation learning process.
% We deem edge embeddings instrumental for capturing critical transition relationships in routing problems, especially those with energy or time constraints.
We consider edge embeddings crucial for capturing important transition relationships in routing problems, particularly those involving energy or time constraints.
Specifically, the \ac{mha} mechanism uses $h=8$ heads, with the last dimension of $\RM{Q},\RM{K},\RM{V}$ being $d_{\rm q}=d_{\rm k}=d_{\rm v}=16$. 
After the attention layers, we apply a node-wise \ac{ff} layer \cite{VasShaParUszJonGomKaiPol:C17}. The \ac{ff} layer is composed of two linear layers and a ReLU activation with dimension $d_{\rm ff}=512$. 
Residual connections \cite{HeZhaRen:C16} and instance normalization \cite{UlyVedLem:A16} are applied to both the integrated attention layer and the \ac{ff} layer.
% The final graph embeddings $\{\RV{n}_{i}^{L}\}_{i=1,2,\ldots,N}$ and $\{\RV{e}_{i,j}^{L}\}_{i,j=1,2,\ldots,N}$ are generated after $L$ encoder layers.
Finally, the graph embeddings $\{\RV{n}_{i}^{L}\}_{i=1,2,\ldots,N}$ and $\{\RV{e}_{i,j}^{L}\}_{i,j=1,2,\ldots,N}$ are generated after $L$ encoder layers.

The decoder leverages the graph embeddings and context to generate the probability distribution for the next node.
The context embedding layer is responsible for encoding the context information at the current time step $t$.
% The context embedding layer is responsible for encoding the information required for decision-making at time step $t$. 
The composition of context depends on the specific problem setting.
For example, in the \ac{tsp}, the context includes the embeddings of the current node and the starting node:
\begin{equation}
\setlength{\abovedisplayskip}{5pt}
\setlength{\belowdisplayskip}{5pt}
\hat{\RV{n}}_{\rm c}=\operatorname{Concat}(\RV{n}_{\pi_{t-1}}^L,\RV{n}_{\pi_1}^L).
\label{cetsp}
\end{equation}
In the case of the \ac{cvrp}, where the starting node is fixed, the context includes the current node embedding along with the remaining vehicle capacity:
\begin{equation}
\setlength{\abovedisplayskip}{5pt}
\setlength{\belowdisplayskip}{5pt}
\hat{\RV{n}}_{\rm c}=\operatorname{Concat}(\RV{n}_{\pi_{t-1}}^L,\operatorname{Cap}(\V{\pi}_{<t},\RV{s})).
\label{cecvrp}
\end{equation}
We provide further context details on different routing problems in Appendix \ref{appendixsec:problem processing}. 
A linear Layer is used to project the context vector into the embedding space:
\begin{equation}
\setlength{\abovedisplayskip}{5pt}
\setlength{\belowdisplayskip}{5pt}
{\RV{n}}_{\rm c}^0=\M{W}_{\rm c}\hat{\RV{n}}_{\rm c}+\V{b}_{\rm c}.
\label{contextembed}
\end{equation}

The attention mechanism is also incorporated in our decoder for decision-making.
First, the context embedding is updated through integrated attention layers, which consist of Node-Node and Node-Edge attention.
These layers enable ${\RV{n}}_{\rm c}^0$ to gather information from adjacent nodes $\{\RV{n}_{i}^{L}\}_{i=1,2,\ldots,N}$ and edges $\{\RV{e}_{\pi_{t-1},j}^{L}\}_{j=1,2,\ldots,N}$. 
% Specifically, Node-Node and Node-Edge attention are employed to enable ${\RV{n}}_{\rm c}^0$ to perceive information from adjacent nodes $\{\RV{n}_{i}^{L}\}_{i=1,2,\ldots,N}$ and edges $\{\RV{e}_{\pi_{t-1},j}^{L}\}_{j=1,2,\ldots,N}$. 
% In this part, we apply masks to \cref{nna,nea} to ensure that only feasible nodes are considered.
To ensure that only feasible nodes are considered, we apply masks to the attention operations.
The final probability is computed via a single-head attention layer. 
We perform a dot-product attention operation between the context embedding and adjacent edge embeddings. 
The query and keys are projected using parameter matrices $\M{W}^{\rm q}_{\rm out},\M{W}^{\rm k}_{\rm out}\in\mathbb{R}^{d_{\rm m}\times d_{\rm m}}$:
\begin{equation}
\setlength{\abovedisplayskip}{5pt}
\setlength{\belowdisplayskip}{5pt}
{\RV{q}}_{\rm c}=\M{W}^{\rm q}_{\rm out}{\RV{n}}_{\rm c}^K,\quad 
{\RV{k}}_{j}=\M{W}^{\rm k}_{\rm out}\RV{e}_{\pi_{t-1},j}^{L}.
\label{outpre}
\end{equation}
% Then we apply a clipped tanh function to limit the range of attention weights, followed by a softmax function to compute the final probabilities:
We then apply a clipped tanh function to restrict the range of the attention weights, followed by a softmax function to compute the final probabilities:
{\setlength{\abovedisplayskip}{5pt}
\setlength{\belowdisplayskip}{5pt}
\begin{align}
&u_j=\begin{cases}
\alpha\cdot\operatorname{tanh}\left(\frac{{\RV{q}}_{\rm c}^{\rm T}{\RV{k}}_{j}}{\sqrt{d_{\rm m}}}\right),&\text{if node $j$ is feasible}\\
-\infty,&\text{otherwise}
\end{cases}\\
&p_{\V{\theta}}(\pi_t=n_i|\V{\pi}_{<t},\RV{s})=\operatorname{softmax}_i(u_1,u_2,\ldots,u_N).
\end{align}}%

% decoding type
% appendix
The node sequence is decoded from $p_{\V{\theta}}(\pi_t|\V{\pi}_{<t},\RV{s})$ using two strategies: sampling and greedy. 
During training, we adopt the sampling strategy to encourage exploration within the feasible region.
During validation and testing, the greedy strategy selects the node with the highest probability at each step to generate a single trajectory. In contrast, the sampling strategy randomly selects nodes based on probabilities, producing multiple results at the cost of increased computation time.

\subsection{Reinforcement learning with symmetry-based baseline}
We train the proposed model using the REINFORCE algorithm \cite{Wil:J92} with a symmetry-based baseline.
REINFORCE is a classic policy gradient algorithm that estimates the gradient by evaluating the cost of an entire trajectory.
The loss function is defined as $\mathcal{L}(\V{\theta}|\RV{s})=\mathbb{E}_{p_{\V{\theta}}(\V{\pi}|\RV{s})}[C(\V{\pi}, \RV{s})]$, where $p_{\V{\theta}}$ represents the probability of generating a solution $\V{\pi}$ for a given instance $\RV{s}$:
\begin{equation}
\setlength{\abovedisplayskip}{5pt}
\setlength{\belowdisplayskip}{5pt}
p_{\V{\theta}}(\V{\pi}|\RV{s})=p({\pi}_1)\prod_{t=2}^{T}p_{\V{\theta}}(\pi_t|\V{\pi}_{<t},\RV{s}).
\label{ppi}
\end{equation}
The REINFORCE gradient estimator is then expressed as:
\begin{equation}
\setlength{\abovedisplayskip}{5pt}
\setlength{\belowdisplayskip}{5pt}
\nabla\mathcal{L}(\V{\theta}|\RV{s})=\mathbb{E}_{p_{\V{\theta}}(\V{\pi}|\RV{s})}\left[\left(C(\V{\pi}, \RV{s})-b(\RV{s})\right)\nabla\log  p_{\V{\theta}}(\V{\pi}|\RV{s})\right]
\label{rlloss}
\end{equation}
The baseline function $b(\RV{s})$ is introduced to provide a reference cost for each instance. 
A well-designed baseline can significantly improve training efficiency and accelerate convergence.
\cite{SutBar:B98,KimParPar:C22}.

% baseline
We adopt a baseline function derived from the symmetries in routing problems \cite{KimParPar:C22}, including starting point symmetry, sampling symmetry, and graph symmetry.
Starting point symmetry applies to problems like the \ac{tsp}, where no fixed starting point is specified, suggesting that the optimal solution should remain invariant regardless of the selected starting node \cite{KwoChoKimYooGwoMin:C20}.
Sampling symmetry asserts that the expected cost should remain consistent across multiple trajectories sampled from the model's output for a given instance.
Finally, graph symmetry concerns the relative topological relationships within the graph structure, implying that the model should produce similar solution sequences for instances that are equivalent under transformations like rotation or reflection, as the underlying problem remains unchanged \cite{KimParPar:C22}.
% Starting point symmetry addresses problems like the \ac{tsp} where no fixed starting point is defined, positing that the optimal solution should be invariant regardless of the chosen starting node \cite{KwoChoKimYooGwoMin:C20}. 
% Sampling symmetry asserts that the expected cost should be consistent across multiple trajectories sampled from the model's output with a given instance. 
% Finally, graph symmetry concerns the relative topological relationships within the graph structure; it implies that the model should output similar solution sequences for instances that are equivalent under transformations such as rotation or reflection, as the underlying problem remains unchanged \cite{KimParPar:C22}.

% encoder,训练的效率
% 几何等价类
Computationally, we combine starting point and sampling symmetry. 
For \ac{tsp}, the starting point is sampled from a uniform distribution, i.e.,$p(\pi_1 = n_i) = 1/N$. 
For a given instance $\RV{s}$, we generate $m$ equivalent problems by applying random rotations and reflections. 
The model then produces $n$ solutions for each transformed instance under its sampling policy.
The baseline $b(\RV{s})$ is estimated as the mean cost across the $m \times n$ solutions:
\begin{equation}
\setlength{\abovedisplayskip}{5pt}
\setlength{\belowdisplayskip}{5pt}
b(\RV{s}) \approx \frac{1}{mn}\sum_{i=1}^m\sum_{j=1}^nC(\V{\pi}^{i,j}, \RV{s}^i),
\end{equation}
where $\{\RV{s}^i\}$ are equivalent problems of $\RV{s}$, and $\{\V{\pi}^{i,j}\}$ are the solutions sampled for $\RV{s}^i$.Therefore, the gradient of $\mathcal{L}(\V{\theta}|\RV{s})$ is estimated by:
\begin{equation}
\setlength{\abovedisplayskip}{5pt}
\setlength{\belowdisplayskip}{5pt}
\nabla\mathcal{L}(\V{\theta}|\RV{s}) \approx \frac{1}{mn}\sum_{i=1}^m\sum_{j=1}^n\left[\left(
C(\V{\pi}^{i,j}, \RV{s}^i)- b(\RV{s})
\right)\nabla\log p_{\V{\theta}}\right].
\label{finalgra}
\end{equation}

% 总结段
% By leveraging the aforementioned symmetries, our baseline function effectively performs data augmentation on the input instances, thereby improving sample efficiency. 
By leveraging the symmetries described above, our baseline function effectively augments the input instances, enhancing sample utilization.
The designed $b(\RV{s})$ establishes a relative benchmark for evaluating the cost of the model's outputs. 
Experimental results demonstrate that this baseline construction method achieves better convergence performance compared to the deterministic greedy rollout approach \cite{KoovanWel:C19, KimParPar:C22}.
% leads to superior convergence performance compared to the deterministic greedy rollout approach \cite{KoovanWel:C19}.
Based on the gradient expression in \cref{finalgra}, we employ the Adam optimizer \cite{KinBa:C15} to learn the model parameters.

\begin{table*}[ht!]
\caption{Experimental results on \ac{tsp}, \ac{cvrp} and PCTSP. The downward arrows in the metrics indicate that smaller values are better. A '/' denotes that the method is not applicable or has not been applied to the corresponding problem. The optimal methods under the greedy and sampling strategies for learning-based approaches are highlighted in bold.}
\label{table:results of tsp/cvrp/pctsp}
\vskip 0.15in
\begin{center}
\begin{small}
\begin{tabular}{lc|ccc|ccc|ccc}
\toprule
\multirow{2}{*}{Method} & \multirow{2}{*}{Type} & \multicolumn{3}{c|}{{TSP}} & \multicolumn{3}{c|}{CVRP} & \multicolumn{3}{c}{PCTSP}\\
% \cmidrule(lr){3-4} \cmidrule(lr){5-6}
 & & Cost$\downarrow$ & Gap$\downarrow$ & Time$\downarrow$ & Cost$\downarrow$ & Gap$\downarrow$& Time$\downarrow$ & Cost$\downarrow$ & Gap$\downarrow$& Time$\downarrow$\\
\midrule
LKH-3 &  Heuristic & 5.70 & 0.00\% & 1.3m & 10.38 & 0.00\% & 57m & / & / & / \\
Gurobi &  Branch and Bound & 5.70 & 0.00\% & 30m & / & / & / & 4.47 & 0.00\% & 58m\\
OR-Tools (1m) &  Heuristic & 5.70 & 0.00\% & 2.6h & 10.56 & 1.78\% & 2.6h & 4.48 & 0.07\% & 2.6h \\
OR-Tools (10m) &  Heuristic & 5.70 & 0.00\% & 1.1d & 10.42 & 0.42\% & 1.1d& 4.47 & 0.00\% & 1.1d  \\
\midrule
GAM &  Greedy & 5.80 & 1.76\% & 4s & 10.98 & 5.86\% & 4s & 4.60 & 2.84\% & 3s \\
GATv2 &Greedy& 5.77 &1.33\% & 3s&10.90 &5.04\% &3s &4.56 &1.95\% &2s\\
% GREAT &Greedy& & & & & & & & &\\
POMO & Greedy & 5.73 & 0.64\% & 5s & 10.74 & 3.54\% & 6s & / & / & / \\
Sym-NCO & Greedy & 5.73 & 0.57\% & 4s & 10.73 & 3.42\% & 4s & 4.52 & 1.07\% & 4s \\ 
EGAM (\textbf{Ours}) & Greedy & \textbf{5.72} & \textbf{0.49\%} & 6s & \textbf{10.72} & \textbf{3.29\%} & 7s & \textbf{4.51} & \textbf{0.81\%} & 6s  \\
\midrule
GAM &  1280 Sampling & 5.73 & 0.52\% & 5.4m & 10.62 & 2.40\% & 6.1m & 4.52 & 1.05\% & 4.2m \\
GATv2 &1280 Sampling&5.72 & 0.33\%&3.2m &10.59 &2.05\% &3.8m &4.50 &0.55\% &2.5m\\
% GREAT &Sampling& & & & & & & & &\\
POMO & 8$\times$20 Sampling & 5.70 & 0.06\% & 2.7m & 10.47 & 0.98\% & 2.9m & / & / & / \\ 
Sym-NCO & 8$\times$20 Sampling & 5.70 & 0.04\% & 1.4m & \textbf{10.47} & \textbf{0.94\%} & 1.6m & 4.48 & 0.25\% & 1.3m \\ 
EGAM (\textbf{Ours}) & 8$\times$20 Sampling & \textbf{5.70} & \textbf{0.03\%} & 2.2m & 10.48 & 1.01\% & 2.4m & \textbf{4.48} & \textbf{0.11\%} & 2.3m \\
\bottomrule
\end{tabular}
% \end{sc}
\end{small}
\end{center}
% \vskip -0.1in
\end{table*}

\begin{table*}[ht!]
\caption{Experimental results on \ac{tsptw} and \ac{tspdl}. Notations are the same with \cref{table:results of tsp/cvrp/pctsp}.}
\label{table:results of tsptw/tspdl}
\vskip 0.15in
\begin{center}
\begin{small}
\begin{tabular}{lc|cccc|cccc}
\toprule
\multirow{2}{*}{Method} & \multirow{2}{*}{Type} & \multicolumn{4}{c|}{TSPTW} & \multicolumn{4}{c}{{TSPDL}}  \\
% \cmidrule(lr){3-4} \cmidrule(lr){5-6}
 & & Cost$\downarrow$ & Infeasible$\downarrow$ & Gap$\downarrow$ & Time$\downarrow$ & Cost$\downarrow$ & Infeasible$\downarrow$ & Gap$\downarrow$& Time$\downarrow$ \\
\midrule
LKH-3 &  Heuristic & 13.01 & 0.00\% & 0.00\% & 1.1h & 10.86 & 0.00\% & 0.00\% & 1.4h \\
% OR-Tools (1m) &  Heuristic  & 13.39 & 15.52\% & 2.90\% & 2.2h& ? & ? & ? & ?\\
OR-Tools (10m) &  Heuristic & 13.05 & 15.52\% & 0.30\% & 22h & / & / & / & / \\
% Greedy-C &  Heuristic & 25.33 & 47.52\% & 94.70\% & 4s & 26.09  & 0.00\% & ? & 4s\\
\midrule
GAM &  Greedy & 14.58 & 20.86\% & 12.06\% & 4s & 12.43 & 10.27\% & 14.41\% & 4s\\
GATv2 &Greedy&14.48 & 18.60\% &11.32\% &3s & 11.97&10.43\% &10.25\% & 3s\\
% GREAT &Greedy& & & & & & & & \\
POMO & Greedy & 14.07 & 14.92\% & 8.13\% & 5s & 11.69 & 7.60\% & 7.67\% & 5s \\
Sym-NCO & Greedy & 14.10 & 17.88\% & 8.39\% & 4s & 11.70 & 3.74\% & 7.71\% & 4s \\ 
EGAM (\textbf{Ours}) & Greedy & \textbf{13.77} & \textbf{7.76\%} & \textbf{5.84\%} & 6s & \textbf{11.45} & \textbf{1.14\%} & \textbf{5.46\%} & 5s \\
\midrule
GAM &  1280 Sampling & 13.90 & 0.74\% & 6.83\% & 5.5m & 11.82 & 1.21\% & 8.86\% & 5.6m  \\
GATv2 &1280 Sampling& 13.66&0.33\% &4.97\% &3.2m &11.38 &0.10\% &4.75\% &3.1m \\
% GREAT &Sampling& & & & & & & & \\
POMO & 8$\times$20 Sampling & 13.54 & 0.68\% & 4.06\% & 2.8m & 11.32 & 1.46\% & 4.26\% & 3.0m \\ 
Sym-NCO & 8$\times$20 Sampling & 13.49 & 0.37\% & 3.69\% & 1.5m & 11.30 & 0.51\% & 4.03\% & 1.5m \\ 
EGAM (\textbf{Ours}) & 8$\times$20 Sampling & \textbf{13.42} & \textbf{0.31\%} & \textbf{3.19\%} & 2.2m & \textbf{11.19} & \textbf{0.09\%} & \textbf{3.02\%} & 2.2m \\
\bottomrule
\end{tabular}
% \end{sc}
\end{small}
\end{center}
\vskip -0.1in
\end{table*}

\subsection{Scalability}
% 基础模型、
In the previous sections, we established the autoregressive structure of \ac{egam} and the reinforcement learning training framework. 
Notably, our model also retains the ability to be extended into a non-autoregressive mode. 
By preserving the Encoder for embedding the graph structure, a modified decoder can be employed to output routing planning heatmaps. 
Furthermore, the modified model can be trained using \ac{sl} \cite{SunYan:C23,LuoLinLiu:C23}. 
The proposed framework is also adaptable to additional methods aimed at assisting decision-making in graph neural networks \cite{FuQiuZha:C21,BiMaZho:C24,XiaYanLiu:C24}, thereby enhancing generalization across diverse problem settings and scales.
% The proposed framework is also adaptable for additional methods aimed at assisting decision-making in graph neural networks \cite{FuQiuZha:C21,BiMaZho:C24,XiaYanLiu:C24}, thereby enhancing generalization across various problem settings and scales.
% 非自回归、监督训练

\section{Experiments}
% Experiment setting
% According to the hardness of the constraints, the experiments were divided into two groups: (1) \ac{tsp}, \ac{cvrp}, \ac{pctsp}, and (2) \ac{tsptw}, \ac{tspdl}, \ac{vrptw}.
% 添加两类问题的说明
We evaluated the proposed model on several routing problems, setting the number of nodes to 50. 
Based on the complexity of the constraints, the experiments were divided into two groups:
(1) \ac{tsp}, \ac{cvrp}, and \ac{pctsp}, which feature relatively simple constraints and minimal restrictions on the node sequence;
and (2) \ac{tsptw}, \ac{tspdl}, and \ac{vrptw}, which involve more challenging constraints, such as time windows and draft limits, making solution feasibility highly sensitive to the ordering of nodes.
% According to the hardness of the constraints, the experiments were divided into two groups: (1) \ac{tsp}, \ac{cvrp}, and \ac{pctsp}, where constraints are relatively simple and impose minimal restrictions on the node sequence; and (2) \ac{tsptw}, \ac{tspdl}, and \ac{vrptw}, which involve hard constraints such as time windows and draft limits, making solution feasibility highly sensitive to the ordering of nodes
During the model training phase, the experiments were conducted on four NVIDIA RTX 3090 (24GB) GPUs. 
Our model uses four encoder layers and one decoder layer.
The batch size was set to 128, with each epoch containing 2500 batches. 
The model was trained for 100 epochs, with the learning rate $\eta=10^{-4}$ for the first 90 epochs and $\eta=10^{-5}$ for the last 10 epochs.
% During the testing phase, only one GPU is used to ensure a fair comparison of time complexity with traditional search-based methods. 
For the testing phase, a single GPU was used to ensure a fair comparison of time complexity with traditional search-based methods.

We considered two categories of comparative algorithms: 
(1) Traditional routing problem solvers, including LKH-3\footnote{\url{http://webhotel4.ruc.dk/~keld/research/LKH-3/}} \cite{Hel:T17}, Gurobi\footnote{\url{https://www.gurobi.com/solutions/gurobi-optimizer/}} (Gurobi Optimization), and OR-Tools\footnote{\url{https://developers.google.com/optimization/routing/}} (Google). 
The performance of these solvers was validated on a CPU with 64 cores. 
LKH-3 is a well-established heuristic solver, often regarded as capable of obtaining almost optimal solutions for problems where it is applicable.
Gurobi employs the branch-and-bound method to find exact solutions to combinatorial optimization problems. 
OR-Tools is another widely used solver, where the quality of the results depends on the search time. We evaluated OR-Tools' performance under different time limits to provide a comprehensive reference.
% LKH-3 is a well-established heuristic solver, often regarded as capable of obtaining near-optimal solutions for various routing problems. 
% LKH-3 is a mature heuristic solver for various routing problems, often regarded as capable of obtaining optimal solutions. 
% Gurobi uses the branch-and-bound method to find exact solutions for combinatorial optimization problems. 
% OR-Tools is another commonly used tool, where the quality of results is dependent on the search time. Thus, we evaluated OR-Tools’ performance under different time limits to provide a reference. 
(2) Learning-based methods, including GAM \cite{KoovanWel:C19,BiMaZho:C24}, GATv2 \cite{BroAloYah:C22}, POMO \cite{KwoChoKimYooGwoMin:C20,BiMaZho:C24}, and Sym-NCO \cite{KimParPar:C22}. 
We evaluated both greedy and sampling inference strategies during the testing phase. The greedy strategy generates a single trajectory, directly reflecting the performance of NCO solvers. The sampling strategy involves input augmentation and repeated experiments to select the best trajectory. 
For different methods, we employed two sampling modes: repeating the sampling process 1280 times, or using input augmentation 8 times followed by 20 repetitions of the decision process.
% We considered both greedy and sampling inference strategies during the testing phase. The greedy strategy generates a single trajectory, offering a direct reflection of performances of NCO solvers. 
% The sampling strategy involves input augmentation and repeated experiments to select the best trajectory. 
% For different methods, we employed two sampling modes: repeating the sampling process 1280 times, or using input augmentation 8 times followed by 20 repetitions of the procedure.
% The sampling strategy involves input augmentation and repeated experiments to select the best trajectory.
% The detailed experimental results are as follows.
% The computational resources for training and testing were kept consistent with the proposed model.

\subsection{\ac{tsp} \& \ac{cvrp} \& \ac{pctsp}}

We selected \ac{tsp}, \ac{cvrp} and \ac{pctsp} as representatives of routing problems with relatively simple constraints. The TSP aims to find the shortest route that visits each location exactly once and returns to the origin. 
The \ac{cvrp} extends the \ac{tsp} by introducing vehicle capacity constraints, requiring routes to serve customer nodes without exceeding vehicle capacity. 
% Vehicles must return to the depot for reloading when necessary, with the objective of minimizing total travel distance.
Vehicles must return to the depot to unload when their capacity is reached, with the objective of minimizing the total travel distance.
The \ac{pctsp} builds upon the \ac{tsp} by associating each node with a prize and a penalty, aiming to collect prizes from visited nodes while minimizing penalties for skipped nodes. A feasible tour ensures the total prize meets a predetermined threshold, with the overall cost being the sum of the tour length and penalties for omitted nodes. 
Further details on these problems are provided in the Appendix \ref{appendixsec:problem processing}.
Our testing set includes 10,000 randomly generated instances for each problem, and the large-scale dataset helps mitigate performance variations caused by randomness.

% 解释为什么cvrp采样没有sota好。
% We evaluated the proposed model and baseline algorithms on the same datasets comprising 10,000 instances, comparing their average cost, optimality gap and total solving time. 
% The experimental results on \ac{tsp}, \ac{cvrp}, and \ac{pctsp} are presented in \cref{table:results of tsp/cvrp/pctsp}. 
% As can be observed, the proposed algorithm consistently outperforms other \ac{nco} solvers under the greedy strategy. 
% Especially, \ac{egam} achieves a slight improvement over state-of-the-art approaches POMO and Sym-NCO.
% Under the sampling strategy, EGAM achieves performance that matches POMO and Sym-NCO, while demonstrating a clear advantage over GAM and GATv2.
% In terms of solution speed, EGAM is close to other NCO solvers, capable of completing greedy inference for 10,000 instances within a few seconds. 
% While LKH-3 and Gurobi can obtain exact solutions, they require significantly more time. 
% In contrast, the proposed method can generate reasonably good solutions within a relatively short timeframe and can sample for better results when more time is available.
The experimental results on the \ac{tsp}, \ac{cvrp}, and \ac{pctsp} are presented in \cref{table:results of tsp/cvrp/pctsp}. 
We report the average cost, average optimality gap, and total computation time for each method across each problem's testing set. 
As can be observed, our model consistently outperforms other NCO solvers under the greedy strategy. 
Notably, \ac{egam} shows a slight advantage over the state-of-the-art approaches POMO and Sym-NCO. Under the sampling strategy, \ac{egam}, POMO, and Sym-NCO perform almost identically, while \ac{egam} maintains a clear advantage over GAM and GATv2. 
In the case of \ac{cvrp} under the sampling strategy, \ac{egam} performs slightly worse than Sym-NCO, but the difference is only 0.07\%, which we attribute to the training process focusing on optimizing single-decision performance. 
In summary, for these three problems with simple constraints, \ac{egam} either matches or slightly outperforms state-of-the-art methods.

\begin{table}[t]
\caption{Experimental results on \ac{vrptw}. Notations are the same with \cref{table:results of tsp/cvrp/pctsp} and \cref{table:results of tsptw/tspdl}. }
\label{table:results of vrptw}
\vskip 0.in
\begin{center}
\begin{small}
\begin{tabular}{l|ccc}
\toprule
\multirow{2}{*}{Method} & \multicolumn{3}{c}{{VRPTW}}\\
% \cmidrule(lr){3-4} \cmidrule(lr){5-6}
 &  Cost$\downarrow$ & Gap$\downarrow$ & Time$\downarrow$\\
\midrule
OR-Tools (1m)  & 20.03 & 8.91\% & 2.6h  \\
OR-Tools (10m) & 19.07 & 3.70\% & 1.1d \\
OR-Tools (60m) & 18.39 & 0.00\% & 6.4d \\
\midrule
GAM (Greedy) & 21.46 & 16.71\% & 3s  \\
GATv2 (Greedy)&20.48 &11.36\% &2s  \\
% GREAT  (Greedy) & &  \\
Sym-NCO (Greedy) & 19.97 & 8.62\% & 3s  \\ 
EGAM (\textbf{Ours}, Greedy) & \textbf{19.60} & \textbf{6.58\%} & 5s  \\
\midrule
GAM (1280 Sampling) & 19.47 & 5.90\% & 3.9m \\
GATv2 (1280 Sampling)&19.12 &3.97\% &2.5m  \\
% GREAT (Sampling)& & &  \\
Sym-NCO (8$\times$20 Sampling) & 18.96 & 3.12\% & 1.2m  \\ 
EGAM (\textbf{Ours}, 8$\times$20 Sampling) & \textbf{18.83} & \textbf{2.45\%} & 1.9m \\
\bottomrule
\end{tabular}
% \end{sc}
\end{small}
\end{center}
\vskip -0.1in
\end{table}

\subsection{\ac{tsptw} \& \ac{tspdl} \& \ac{vrptw}}

\ac{tsptw} introduces time window constraints on the nodes to be visited. 
Each node must be visited within its designated time window; otherwise, the visit is considered a failure. 
% If the agent arrives before the time window opens, it must wait until the window becomes available. 
If arrival occurs earlier than the start of the time window, the agent must wait until the window opens. 
A tour is feasible only if all nodes are visited while adhering to these time constraints. 
% Of course, d
During data generation, we ensure that feasible solutions exist.
For learning-based methods, we incorporate a penalty term for time window violations, guiding the model to focus on learning feasible solutions. 
\ac{tspdl} involves ship routing, where each port (node) has a cargo capacity demand and a draft limit. The goal is to plan a route such that the ship's load at each port does not exceed its draft limit. Similarly, we introduce penalties for violations of the draft limit.
For \ac{vrptw}, we relax the constraint of visiting all nodes. The objective is modified to visit as many nodes as possible within their time windows. During data generation, we allow time window conflicts, which aligns more closely with real-world scenarios. The cost function is set to the number of nodes that were not successfully visited.
% These three problems, due to the presence of time windows or draft limits, impose higher demands on neural network decision-making.
In addition to the previously mentioned metrics, we also report the infeasibility rates for TSPTW and TSPDL. Each problem's testing set consists of 10,000 instances.

The experimental results on \ac{tsptw} and \ac{tspdl} are presented in \cref{table:results of tsptw/tspdl}.
The results demonstrate that \ac{egam} significantly outperforms other \ac{nco} methods in both cost and infeasible rate. 
Specifically, under a single-decision (greedy) setting, \ac{egam} reduces the optimality gap by 2.29\% and the infeasible rate by 7.16\% compared to POMO on \ac{tsptw}. 
For \ac{tspdl}, \ac{egam} also achieves reductions of 2.21\% in optimality gap and 2.60\% in infeasible rate compared to the strongest baselines. 
When employing the sampling strategy, \ac{egam} remains superior to all comparative algorithms. 
Notably, while achieving lower costs, it reduces the infeasible rates on \ac{tsptw} and \ac{tspdl} to 0.31\% and 0.09\%, respectively.
The experimental results on \ac{vrptw} are presented in \cref{table:results of vrptw}. 
Since the problem formulation we consider falls outside the solvable scope of LKH-3, we use OR-Tools with a search time limit of one hour to provide an approximate reference for optimal values. 
Compared to Sym-NCO, \ac{egam} reduces the cost by 2.04\% under the greedy strategy and by 0.67\% under the sampling strategy. 
Furthermore, the sampling results from \ac{egam} surpass those obtained by OR-Tools with a 10-minute limit, while being approximately 800 times faster.
% 提附录以及代码开源

In Appendix \ref{appendix:addotional experiments}, we provide additional experiments and analyses, including training dynamics, the impact of model size, and generalization tests. 
% The model implementation and experimental code are included in the supplementary material. 
We intend to release our source code publicly to support reproducibility and foster collaboration within the research community in the future.

\section{Conclusion}
In this paper, we proposed a novel neural network framework, \ac{egam}, designed for routing problems.
We introduced generalized graph attention mechanism and applied it to graph embedding and autoregressive decision-making. 
The model was trained using \ac{rl} with a symmetry-based baseline, requiring no expert knowledge or labeled data. 
Experimental results demonstrated that \ac{egam} matched or surpassed existing \ac{rl}-based NCO solvers across a variety of routing problems. 
Notably, \ac{egam} exhibits particularly strong performance on highly constrained problems such as \ac{tsptw}, \ac{tspdl}, and \ac{vrptw}.
Our future work will focus on the scalability of \ac{egam}, including the following directions: 
(1) integrating improved methods tailored for \ac{nco} to enhance performance on specific problem types and larger-scale instances; 
(2) extending its application to a wider variety of routing problems, such as edge-based route planning; 
(3) exploring additional training paradigms, such as \ac{sl}; and 
(4) generalizing the approach beyond routing problems to tackle a wider variety of combinatorial optimization tasks, with the goal of establishing a universal framework for graph-structured problems.
% (4) generalizing from routing problems to broader combinatorial optimization tasks, to establish a universal framework for processing graph-structured problems.

% \section{Acknowledgements}
% \textbf{Do not} include acknowledgements in the initial version of the paper submitted for blind review.

% If a paper is accepted, the final camera-ready version can (and usually should)
% include acknowledgements.  Such acknowledgements should be placed at the end of
% the section, in an unnumbered section that does not count towards the paper
% page limit. Typically, this will include thanks to reviewers who gave useful
% comments, to colleagues who contributed to the ideas, and to funding agencies
% and corporate sponsors that provided financial support.

\section*{Impact Statement}
This paper presents work whose goal is to advance the field of 
Machine Learning. There are many potential societal consequences 
of our work, none which we feel must be specifically highlighted here.

% \nocite{BahChoBen:A14}

\nocite{LuoLinLiu:C23}
\nocite{LuoLinWan:A24}
\nocite{XiaZhaChe:C24}

\bibliography{example_paper}
\bibliographystyle{icml2026}

%%%%%%%%%%%%%%%%%%%%%%%%%%%%%%%%%%%%%%%%%%%%%%%%%%%%%%%%%%%%%%%%%%%%%%%%%%%%%%%
%%%%%%%%%%%%%%%%%%%%%%%%%%%%%%%%%%%%%%%%%%%%%%%%%%%%%%%%%%%%%%%%%%%%%%%%%%%%%%%
% APPENDIX
%%%%%%%%%%%%%%%%%%%%%%%%%%%%%%%%%%%%%%%%%%%%%%%%%%%%%%%%%%%%%%%%%%%%%%%%%%%%%%%
%%%%%%%%%%%%%%%%%%%%%%%%%%%%%%%%%%%%%%%%%%%%%%%%%%%%%%%%%%%%%%%%%%%%%%%%%%%%%%%
\newpage
\appendix
\onecolumn
\section{Details of model architecture}
\label{appendixsec:model details}

\subsection{Encoder}
\label{appendixsubsec:encoder}
% The role of the encoder is to embedding the nodes and edges of the input graph into high-dimensional space.
% For a routing problem instance $\RV{s}$, We represent it using a graph $G=({\Set V},{\Set E})$, while ${\Set V}=\{n_i\mid i=1,2,\ldots,N\}$ and ${\Set E}=\{e_{i,j}\mid i,j=1,2,\ldots,N\}$. The initial node features and edge features are respectively denoted as $\RV{n}_i^{\rm init}\in\mathbb{R}^{F_{\rm n}}$ and $\RV{e}_{i,j}^{\rm init}\in\mathbb{R}^{F_{\rm e}}$. The initial feature dimensions $F_{\rm n}$ and $F_{\rm e}$ depends on problem settings.
The role of the encoder is to embed the nodes and edges of the input graph into a high-dimensional space.
For a routing problem instance $\RV{s}$, we represent it using a graph $G=({\Set V},{\Set E})$, where ${\Set V}=\{n_i\mid i=1,2,\ldots,N\}$ and ${\Set E}=\{e_{i,j}\mid i,j=1,2,\ldots,N\}$. The initial node features and edge features are respectively denoted as $\RV{n}_i^{\rm init}\in\mathbb{R}^{F_{\rm n}}$ and $\RV{e}_{i,j}^{\rm init}\in\mathbb{R}^{F_{\rm e}}$. The initial feature dimensions $F_{\rm n}$ and $F_{\rm e}$ depend on the problem settings.

As shown in \cref{fig:model}, the input features are first transformed into the embedding space $\mathbb{R}^{d_{\rm m}}$ with dimension $d_{\rm m}=128$ through an initial embedding layer:
{\setlength{\abovedisplayskip}{5pt}
\setlength{\belowdisplayskip}{5pt}
\begin{align}
\RV{n}_{i}^{0} = \M{W}_{\rm n}^{\rm init}\RV{n}_{i}^{\rm init}+\M{b}_{\rm n}^{\rm init},\\
\RV{e}_{i,j}^{0} = \M{W}_{\rm e}^{\rm init}\RV{e}_{i,j}^{\rm init}+\M{b}_{\rm e}^{\rm init},
\end{align}}%
where $\M{W}_{\rm n}^{\rm init}\in\mathbb{R}^{d_{\rm m}\times F_{\rm n}},\M{W}_{\rm e}^{\rm init}\in\mathbb{R}^{d_{\rm m}\times F_{\rm e}}, \M{b}_{\rm n}^{\rm init},\M{b}_{\rm e}^{\rm init}\in\mathbb{R}^{d_{\rm m}}$.

% After the initial embedding layer, the node and edge embeddings are updated by multiple encoder layers.
% The encoder layer consists of two components: integrated attention layers and feed-forward layers. The integrated attention layer sequentially performs multi-head attention operations in the order of Node-Node, Edge-Node, and Node-Edge to alternately update the node and edge embeddings. 
% The feed-forward layer is composed of two linear layers and a ReLU activation. The embeddings are first transformed into a higher-dimensional space $d_{\rm ff}=512$ via the input linear layer, then passed through a ReLU activation, and finally restored to $d_{\rm m}$ by the output linear layer.
% Residual connections and normalization are applied to both the integrated attention layer and the feed-forward layer. 
% Regarding the type of normalization layer, our model supports options such as batch, layer, and instance normalization. In the experiments presented in this paper, we employ instance normalization \cite{UlyVedLem:A16}.
% The computation of a single encoder layer proceeds as follows:
After the initial embedding layer, the node and edge embeddings are updated through multiple encoder layers.
The encoder layer consists of two main components: integrated attention layers and feed-forward layers. 
The integrated attention layer sequentially performs multi-head attention operations in the order of Node-Node, Edge-Node, and Node-Edge, enabling alternate updates of node and edge embeddings. 
The feed-forward layer comprises two linear layers with a ReLU activation function. Specifically, the embeddings are first transformed into a higher-dimensional space with dimension $d_{\rm ff}=512$ via the input linear layer, then passed through a ReLU activation, and finally projected back to dimension $d_{\rm m}$ by the output linear layer.
Residual connections and normalization are applied to both the integrated attention layer and the feed-forward layer. 
Our model supports various normalization options, including batch, layer, and instance normalization. In the experiments presented in this paper, we employ instance normalization \cite{UlyVedLem:A16}.
The computation of a single encoder layer proceeds as follows:
{\setlength{\abovedisplayskip}{5pt}
\setlength{\belowdisplayskip}{5pt}
\begin{align}
\hat{\RV{n}}_{i}^{\ell}&=\operatorname{MHA}(\RV{n}_{i}^{\ell-1}, [\RV{n}_1^{\ell-1},\RV{n}_2^{\ell-1},\cdots,\RV{n}_N^{\ell-1}])+\RV{n}_{i}^{\ell-1},\\
\tilde{\RV{e}}_{i,j}^{\ell}&=\operatorname{MHA}(\RV{e}_{ij}^{\ell-1}, [\hat{\RV{n}}_{i}^{\ell},\hat{\RV{n}}_{j}^{\ell}])+\RV{e}_{ij}^{\ell-1},\\
\tilde{\RV{n}}_{i}^{\ell}&=\operatorname{MHA}(\hat{\RV{n}}_{i}^{\ell}, [\tilde{\RV{e}}_{i,1}^{\ell},\tilde{\RV{e}}_{i,2}^{\ell},\cdots,\tilde{\RV{e}}_{i,N}^{\ell}])+\hat{\RV{n}}_{i}^{\ell},\\
\RV{e}_{i,j}^{\ell} &= \operatorname{Norm}(\overline{\operatorname{FF}}(\operatorname{Norm}(\tilde{\RV{e}}_{i,j}^{\ell}))),\\
\RV{n}_{i}^{\ell} &= \operatorname{Norm}(\overline{\operatorname{FF}}(\operatorname{Norm}(\tilde{\RV{n}}_{i}^{\ell}))),
\end{align}}%
% where $\overline{\operatorname{FF}}$ denote the feed-forward layer with residual connection. For input $\RV{x}=\mathbb{R}^{d_{\rm m}}$, $\overline{\operatorname{FF}}$ is formulated as \cref{ffbar}. The parameter dimensions are $\M{W}^{{\rm ff},1}\in\mathbb{R}^{d_{\rm ff}\times d_{\rm m}},\M{W}^{{\rm ff},2}\in\mathbb{R}^{d_{\rm m}\times d_{\rm ff}}, \V{b}^{{\rm ff},1}\in\mathbb{R}^{d_{\rm ff}},\V{b}^{{\rm ff},2}\in\mathbb{R}^{d_{\rm m}}$.
where $\overline{\operatorname{FF}}$ denotes the feed-forward layer with residual connection. For input $\RV{x}\in\mathbb{R}^{d_{\rm m}}$, $\overline{\operatorname{FF}}$ is formulated as \cref{ffbar}. The parameter dimensions are $\M{W}^{{\rm ff},1}\in\mathbb{R}^{d_{\rm ff}\times d_{\rm m}},\M{W}^{{\rm ff},2}\in\mathbb{R}^{d_{\rm m}\times d_{\rm ff}}, \V{b}^{{\rm ff},1}\in\mathbb{R}^{d_{\rm ff}},\V{b}^{{\rm ff},2}\in\mathbb{R}^{d_{\rm m}}$.
%具体的更新公式
\begin{equation}
\setlength{\abovedisplayskip}{5pt}
\setlength{\belowdisplayskip}{5pt}
\hat{\RV{x}} = \M{W}^{{\rm ff},2}\cdot \operatorname{ReLU}(\M{W}^{{\rm ff},1}\RV{x}+\V{b}^{{\rm ff},1})+\V{b}^{{\rm ff},2}+\RV{x}.
\label{ffbar}
\end{equation}
%MHA操作的细节公式-不写了，没必要

%总结结果
% After processing through a total of $L$ encoder layers, the final node embeddings $\{\RV{n}_{i}^{L}\}_{i=1,2,\ldots,N}$ and edge embeddings $\{\RV{e}_{i,j}^{L}\}_{i,j=1,2,\ldots,N}$ are obtained. For a given instance, the encoder is executed only once. 
% Note that the encoder extracts essential information from the input graph-structured data, thereby assisting the decoder in completing the designated task. 
% Furthermore, the encoder structure is generally transferable and can be applied to other graph problems, such as the Maximal Independent Set.
After processing through a total of $L$ encoder layers, the final node embeddings $\{\RV{n}_{i}^{L}\}_{i=1,2,\ldots,N}$ and edge embeddings $\{\RV{e}_{i,j}^{L}\}_{i,j=1,2,\ldots,N}$ are obtained. 
For a given instance, the encoder is executed only once, extracting essential information from the input graph-structured data to assist the decoder in completing the designated task. 
The encoder structure is generally transferable and can be applied to other graph problems, such as the Maximal Independent Set problem.

\subsection{Decoder}
\label{appendixsubsec:decoder}
% context, Markov，自回归，强化学习
% The proposed \ac{egam} employs an autoregressive decoder structure, which outputs the next node's probability distribution $p_{\V{\theta}}(\pi_t|\V{\pi}_{<t},\RV{s})$ based on the historical path $\V{\pi}_{<t}$. We formulate this process using a Markov Decision Process. The state at the current time step is represented by context information and the embeddings generated by the encoder. 
% The embeddings $\{\RV{n}_{i}^{L}\}_{i=1,2,\ldots,N}$ and $\{\RV{e}_{i,j}^{L}\}_{i,j=1,2,\ldots,N}$
% are static and contain information extracted from the problem instance. 
% The context evolves dynamically throughout the decision-making process, with its design depending on specific problems, as detailed in \cref{appendixsec:problem processing}. The initial context vector ${\RV{n}}_{\rm c}^0\in\mathbb{R}^{d_{\rm c}}$ is embedded by a linear context embedding layer:
The proposed \ac{egam} employs an autoregressive decoder structure that outputs the next node's probability distribution $p_{\V{\theta}}(\pi_t|\V{\pi}_{<t},\RV{s})$ based on the historical path $\V{\pi}_{<t}$. 
We formulate this process as a Markov Decision Process, where the state at the current time step is represented by context information and the embeddings generated by the encoder. 
The embeddings $\{\RV{n}_{i}^{L}\}_{i=1,2,\ldots,N}$ and $\{\RV{e}_{i,j}^{L}\}_{i,j=1,2,\ldots,N}$ are static and contain information extracted from the problem instance. 
The context evolves dynamically throughout the decision-making process, with its design depending on the specific problem, as detailed in \cref{appendixsec:problem processing}. 
The initial context vector ${\RV{n}}_{\rm c}^0\in\mathbb{R}^{d_{\rm c}}$ is embedded by a linear context embedding layer:
\begin{equation}
\setlength{\abovedisplayskip}{5pt}
\setlength{\belowdisplayskip}{5pt}
{\RV{n}}_{\rm c}^0=\M{W}_{\rm c}\hat{\RV{n}}_{\rm c}+\V{b}_{\rm c}, \M{W}_{\rm c}\in\mathbb{R}^{d_{\rm m}\times d_{\rm c}}, \V{b}_{\rm c}\in\mathbb{R}^{d_{\rm m}}.
\label{contextembed}
\end{equation}

% Subsequently, similar to the encoder, we update ${\RV{n}}_{\rm c}^0$ using attention layers and feed-forward layers. In the decoder layers, we employ node-node and node-edge attention mechanisms, treating ${\RV{n}}_{\rm c}^k$ (where $k=0,1,ldots,K$) as a node embedding. 
Subsequently, similar to the encoder, we update ${\RV{n}}_{\rm c}^0$ using attention layers and feed-forward layers. 
In the decoder layers, we employ node-node and node-edge attention mechanisms, treating ${\RV{n}}_{\rm c}^k$ (where $k=0,1,\ldots,K$) as a node embedding. 
The specific operations within a single decoder layer are as follows:
{\setlength{\abovedisplayskip}{5pt}
\setlength{\belowdisplayskip}{5pt}
\begin{align}
\hat{\RV{n}}_{\rm c}^k&=\operatorname{MHA}({\RV{n}}_{\rm c}^{k-1}, [\RV{n}_1^{L},\RV{n}_2^{L},\cdots,\RV{n}_N^{L}];\V{m}_{\rm c})+\RV{n}_{\rm c}^{k-1},\label{den2n}\\
\tilde{\RV{n}}_{\rm c}^k&=\operatorname{MHA}(\hat{\RV{n}}_{\rm c}^k, [\RV{e}_{\V{\pi}_{t-1},1}^{L},\RV{e}_{\V{\pi}_{t-1},2}^{L},\cdots,\RV{e}_{\V{\pi}_{t-1},n}^{L}];\V{m}_{\rm c})+\hat{\RV{n}}_{\rm c}^k,\label{den2e}\\
{\RV{n}}_{\rm c}^k &={\operatorname{FF}}(\tilde{\RV{n}}_{\rm c}^k)+\tilde{\RV{n}}_{\rm c}^k.
\end{align}}%
% masked decoder layer, output layer
% As shown in \cref{den2n,den2e}, we incorporate a mask into the MHA operation to control the set of considered nodes. The mask $\V{m}_{\rm c}=[\V{m}_{{\rm c},1},\V{m}_{{\rm c},2},\cdots\V{m}_{{\rm c},N}]$ is a Boolean vector where $\V{m}_{{\rm c},i}=1$ if node $n_i$ is infeasible, and $\V{m}_{{\rm c},i}=0$ otherwise. For example, in TSP, nodes that have already been visited are excluded from consideration, and returning to the starting point is only allowed after all other nodes have been visited, thereby ensuring that the generated path satisfies the problem constraints.
As shown in \cref{den2n,den2e}, we incorporate a mask into the \ac{mha} operation to control the set of considered nodes. 
The mask $\V{m}_{\rm c}=[\V{m}_{{\rm c},1},\V{m}_{{\rm c},2},\ldots,\V{m}_{{\rm c},N}]$ is a Boolean vector where $\V{m}_{{\rm c},i}=1$ if node $n_i$ is infeasible, and $\V{m}_{{\rm c},i}=0$ otherwise. 
This mask also constitutes part of the dynamic context information, evolving as the decoding process progresses.
For example, in the \ac{tsp}, nodes that have already been visited are excluded from consideration, and returning to the starting point is only allowed after all other nodes have been visited, thereby ensuring that the generated path satisfies the problem constraints.

% 啰嗦一点，什么单头多头啊，贪心采样啊，都搞一点，先写个大概就行。
The final probability is computed via a (masked) single-head attention layer. 
We perform a dot-product attention operation between the context embedding ${\RV{n}}_{\rm c}^K$ and adjacent edge embeddings $\{\RV{e}_{\pi_{t-1},j}^{L}\}_{j=1,2,\ldots,N}$. 
The attention weights serve directly as the metric for node selection probabilities, thereby eliminating the need to compute the value vectors.
The query and keys are projected using parameter matrices $\M{W}^{\rm q}_{\rm out},\M{W}^{\rm k}_{\rm out}\in\mathbb{R}^{d_{\rm m}\times d_{\rm m}}$:
\begin{equation}
\setlength{\abovedisplayskip}{5pt}
\setlength{\belowdisplayskip}{5pt}
{\RV{q}}_{\rm c}=\M{W}^{\rm q}_{\rm out}{\RV{n}}_{\rm c}^K,\quad 
{\RV{k}}_{j}=\M{W}^{\rm k}_{\rm out}\RV{e}_{\pi_{t-1},j}^{L}
\label{outpre}
\end{equation}
% Then we apply a clipped tanh function to limit the range of attention weights, followed by a softmax function to compute the final probabilities:
We then apply a clipped tanh function to limit the range of attention weights, followed by a softmax function to compute the final probabilities:
{\setlength{\abovedisplayskip}{5pt}
\setlength{\belowdisplayskip}{5pt}
\begin{align}
&u_j=\begin{cases}
\alpha\cdot\operatorname{tanh}\left(\frac{{\RV{q}}_{\rm c}^{\rm T}{\RV{k}}_{j}}{\sqrt{d_{\rm m}}}\right),&\text{if $\V{m}_{{\rm c},j}=0$}\\
-\infty,&\text{if $\V{m}_{{\rm c},j}=1$}
\end{cases}\\
&p_{\V{\theta}}(\pi_t=n_i|\V{\pi}_{<t},\RV{s})=\operatorname{softmax}_i(u_1,u_2,\ldots,u_N).
\end{align}}%

Nodes are selected from $p_{\V{\theta}}(\pi_t|\V{\pi}_{<t},\RV{s})$ using two decoding strategies: 1) greedy, i.e., $\pi_t=\operatorname{argmax}_{n_i}(p_{\V{\theta}}(\pi_t=n_i|\V{\pi}_{<t},\RV{s}))$; and 2) sampling according to the probability distribution. The decoder runs multiple times until the entire route is planned.

\section{Adaptations to different routing problems}
\label{appendixsec:problem processing}

\subsection{Travelling Salesman Problem (TSP)}
% \textbf{Definition}\quad \ac{tsp} postulates a path planning scenario for a traveling salesman. It investigates how to identify the shortest possible route that visits each given location exactly once and returns to the origin.
% During instance generation, node positions are randomly distributed within a specified region. The total tour length is computed using the Euclidean distance between nodes.
% The initial features include node coordinates and edge lengths.
\textbf{Definition}\quad The \ac{tsp} postulates a path planning scenario for a traveling salesman, aiming to identify the shortest possible route that visits each given location exactly once and returns to the origin.
During instance generation, node positions are randomly distributed within a specified region, and the total tour length is computed using the Euclidean distance between nodes.
The initial features include node coordinates and edge lengths.

\textbf{Context}\quad The context for TSP contains the embeddings of the starting node and the previously visited node (the current node):
\begin{equation}
\setlength{\abovedisplayskip}{5pt}
\setlength{\belowdisplayskip}{5pt}
\hat{\RV{n}}_{\rm c}=\operatorname{Concat}(\RV{n}_{\pi_{t-1}}^L,\RV{n}_{\pi_1}^L).
\end{equation}

\textbf{Mask}\quad The mask for TSP in the decoder depends solely on the visited nodes. As indicated by \cref{mask tsp}, nodes that have already been visited are masked, and returning to the starting node is permitted only after all other nodes have been visited.
\begin{equation}
\setlength{\abovedisplayskip}{5pt}
\setlength{\belowdisplayskip}{5pt}
\V{m}_{{\rm c},i}=\begin{cases}1,&\text{if $n_i$ has been visited}\\0,&\text{otherwise}\end{cases}(n_i\neq\pi_1), 
\V{m}_{{\rm c},\pi_1}=\begin{cases}0,&\text{All nodes have been visited}\\1,&\text{otherwise}\end{cases}.
\label{mask tsp}
\end{equation}

\subsection{Capacitated Vehicle Routing Problem (CVRP)}
\textbf{Definition}\quad The \ac{cvrp} is defined as a path planning problem for vehicles with capacity constraints. It involves planning a set of routes, each starting and ending at a known depot, to serve a set of customer nodes. The core constraint is that the total demand of nodes served on a single route cannot exceed the vehicle's capacity. 
Consequently, a vehicle must return to the depot to be reloaded before this capacity limit is reached. The optimization objective is to minimize the total travel distance across all routes.
Problem instances are constructed by randomly generating node locations and capacity demands. 
The initial features include node coordinates, demands (normalized) and edge lengths.

\textbf{Context}\quad In \ac{cvrp}, where the depot is fixed and specially embedded, the context incorporates the current node embedding and the remaining vehicle capacity:
\begin{equation}
\setlength{\abovedisplayskip}{5pt}
\setlength{\belowdisplayskip}{5pt}
\hat{\RV{n}}_{\rm c}=\operatorname{Concat}(\RV{n}_{\pi_{t-1}}^L,\operatorname{Cap}(\V{\pi}_{<t},\RV{s})).
\end{equation}
The remaining capacity is calculated as $\operatorname{Cap}(\V{\pi}_{<t},\RV{s}))=C_0-\sum_{\tau=1}^{t-1}D_{\RV{s},\pi_\tau}$, while $C_0$ denotes the total capacity and $D_{\RV{s},\pi_\tau}$ denotes the demand of node $\pi_\tau$.

\textbf{Mask}\quad 
% Except for the visited node, nodes whose demand exceeds the remaining capacity of the vehicle are masked. 
Nodes are masked if they have already been visited or if their demand exceeds the remaining capacity of the vehicle.
Additionally, the depot is always accessible.
\begin{equation}
\setlength{\abovedisplayskip}{5pt}
\setlength{\belowdisplayskip}{5pt}
\V{m}_{{\rm c},i}=\begin{cases}1,&\text{if $D_{\RV{s},i}>\operatorname{Cap}(\V{\pi}_{<t},\RV{s})$ or $n_i$ has been visited }\\0,&\text{otherwise}\end{cases}.
\label{mask cvrp}
\end{equation}

\subsection{Prize Collecting Traveling Salesman Problem (PCTSP)}
\textbf{Definition}\quad The \ac{pctsp} extends the TSP by associating each node with a prize and a penalty. Specifically, a prize is collected if the node is visited, while a penalty is incurred if the node is skipped. 
A tour is deemed feasible only if the total prize collected from visited nodes meets or exceeds a predetermined threshold. The overall cost is defined as the sum of the tour length and the total penalties of nodes omitted from the tour. 
PCTSP also involves an additional depot serving as both the start and end point. The initial features include node coordinates, prize (normalized), penalty, and edge lengths.

\textbf{Context}\quad The context for PCTSP includes the current node embedding and the remaining prize to be collected.

\textbf{Mask}\quad The visited nodes are masked. The depot is masked until sufficient prize has been collected.

\subsection{Travelling Salesman Problem with Time Window (TSPTW)}
\textbf{Definition}\quad The \ac{tsptw} imposes time window constraints on the nodes to be visited in the TSP. Each node must be visited before its deadline; otherwise, the visit is considered a failure. 
If arrival occurs earlier than the start of the time window, the agent must wait until the window opens. 
A tour is feasible only if all nodes are visited while satisfying these time constraints. To ensure the feasibility of instances, we adopt a generation function from the literature \cite{BiMaZho:C24}. 
For learning-based methods, a penalty term for time window violations is incorporated into the training cost, formulated as $C = L_{\rm tour} + \beta\cdot(N_{\rm out} + T_{\rm out})$, where $N_{\rm out}$ and $T_{\rm out}$ denote the number of late-visited nodes and the total tardiness duration, respectively. We fix $n_1$ as the starting point and employ a separate initial embedding layer for it. The initial features include node coordinates, start time window, end time window, and edge lengths.

\textbf{Context}\quad The context for TSPTW includes the current node embedding and current time.

\textbf{Mask}\quad For TSPTW, we adopt a mask similar to that used in TSP. However, such a mask cannot guarantee solution feasibility, which is why we incorporate a penalty term into the cost function. In fact, ensuring the feasibility of solutions for TSPTW is a complex issue that has been studied in the work of \cite{BiMaZho:C24}, who proposed the PIP-mask. The PIP-mask can also be easily adapted for use in EGAM. Nevertheless, to compare the raw performance of the models, we did not employ it in the experiments presented in this paper.

\subsection{Traveling Salesman Problem with Draft Limit (TSPDL)} 
\textbf{Definition}\quad The TSPDL considers a ship routing problem for transporting goods between ports. 
Each node to be visited, representing a port, has a cargo capacity demand and a draft limit. 
The goal is to plan a route such that the ship's load at each port does not exceed the port's draft limit; otherwise, the route is considered infeasible. 
The locations, demands, and draft limits of the ports are generated according to specific distributions. 
The cost is defined as $C = L_{\rm tour} + \beta \cdot (N_{\rm out} + D_{\rm out})$, where $N_{\rm out}$ and $D_{\rm out}$ represent the number of overloaded nodes and the total overload amount.
The initial features include node coordinates, capacity demand, draft limit, and edge lengths.

\textbf{Context}\quad The context for TSPDL includes the current node embedding and the current load.

\textbf{Mask}\quad Similar to TSPTW, we use the same mask as TSP.

\subsection{Vehicle Routing Problem with Time Window (VRPTW)} 
\label{appendixsubsec:vrptw}
\textbf{Definition}\quad The classical \ac{vrptw} requires all nodes to be successfully visited. 
However, real-world routing problems often cannot guarantee conflict-free time windows. 
Hence, we relax this constraint by allowing routes to skip some nodes. 
In our formulation, the objective of \ac{vrptw} is to visit as many nodes as possible within their time windows, with the cost function being the number of unsuccessfully visited nodes. The initial features include node coordinates, start time window, end time window, and edge lengths.

\textbf{Context}\quad The context for VRP includes the current node embedding and current time.

\textbf{Mask}\quad For VRPTW, nodes whose time window ends before the current time are masked. Additionally, once the vehicle returns to the depot, the route is considered end.

% 这一节改成以问题作为子标题，下面这三条为主要内容，并且需要在上一节中提及一下
% \subsection{Formulation of routing problems}
% \label{appendixsubsec:Formulation}
% \subsection{State update and context embedding}
% \label{appendixsubsec:context}
% \subsection{Decoding with feasibility masks}
% \label{appendixsubsec:mask}

\section{Additional Experiments and Analysis}
\label{appendix:addotional experiments}

% \subsection{Details of compared algorithms}

% \textbf{GAM}

% \textbf{GATv2}

% \textbf{POMO}

% \textbf{Sym-NCO}

\subsection{Training dynamics}
\begin{figure}[!t]
\centering
\begin{minipage}[b]{.48\textwidth}
  \centering
  \centerline{\includegraphics[width=1.\linewidth]{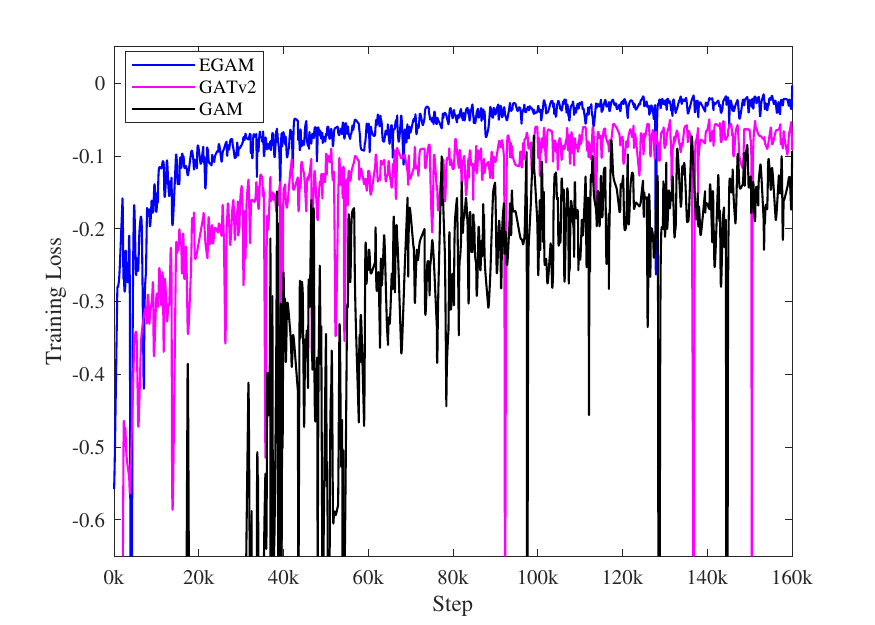}}
  \centerline{(a) Training loss}
  %\medskip
\end{minipage}
\begin{minipage}[b]{.48\textwidth}
  \centering
  \centerline{\includegraphics[width=1.\linewidth]{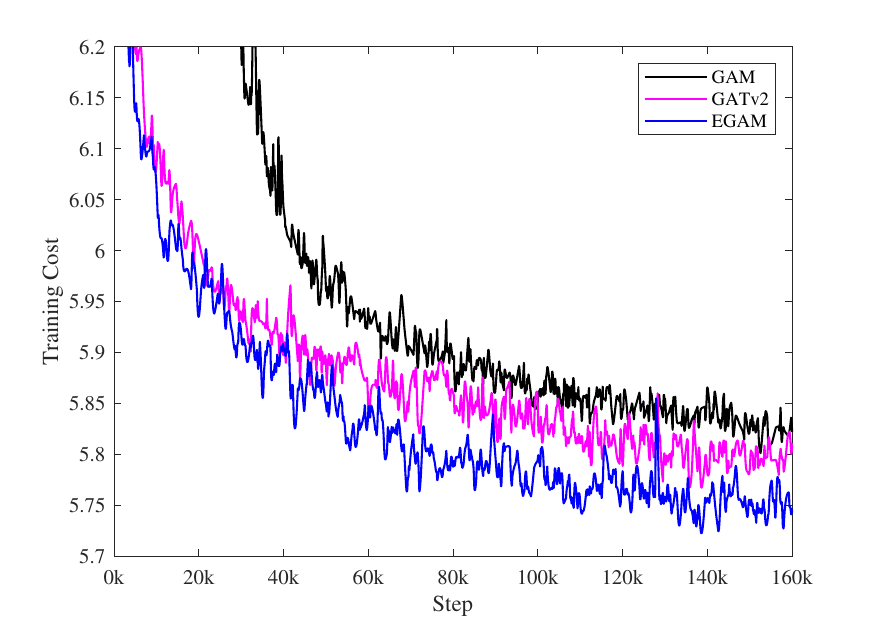}}
  \centerline{(b) Training cost}
  %\medskip
\end{minipage}
% \vspace{-1em}
\caption{Evolution of the training loss and cost for TSP}
\label{fig:td tsp train}
\end{figure}
\begin{figure}[!t]
\begin{minipage}[b]{1.\textwidth}
  \centering
  \centerline{\includegraphics[width=.5\linewidth]{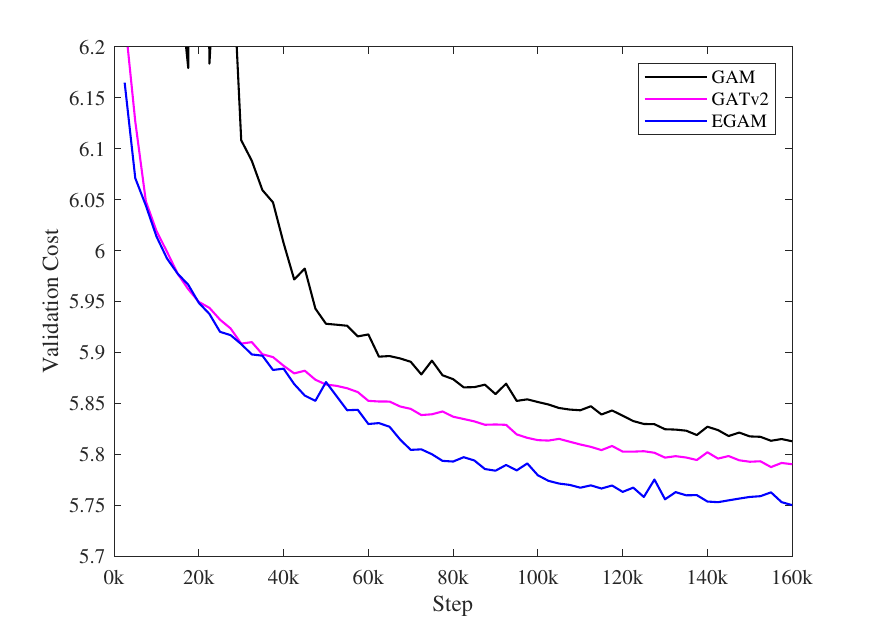}}
%   \centerline{(c) Validation cost}
  %\medskip
\end{minipage}
\vspace{-1em}
\caption{Evolution of the validation cost for TSP}
\label{fig:td tsp val}
\end{figure}
% To comprehensively investigate the training dynamics of different models, we monitor a suite of key metrics during the training process, including the loss, cost and feasible rate (for \ac{tsptw} and \ac{tspdl}) on the training and validation sets. 
% For each routing problem, we train the models for 100 epochs, with each epoch consisting of 2500 batches, where one batch is treated as one training step. 
% The resulting learning curves, presented below, depict the evolution of these metrics for the proposed \ac{egam} and baseline algorithms across different routing problems.

% As shown in \cref{fig:td tsp train} and \cref{fig:td tsp val}, during the training process on TSP, \ac{egam} exhibits a faster decrease in cost on both the training and validation sets compared to \ac{gam} and GATv2. Additionally, the loss of \ac{egam} converges more rapidly toward zero. These results demonstrate the advantages of \ac{egam} in terms of both performance and training speed. 

% As shown in \cref{fig:td tspdl train} and \cref{fig:td tspdl val}.
% Furthermore, for constrained problems such as \ac{tspdl} and \ac{tsptw}, our model effectively balances solution cost and feasibility. As shown in \cref{fig:td tspdl train} and \cref{fig:td tspdl val}, \ac{gam} achieves a significantly higher feasibility rate during training compared to GATv2 and Sym-NCO, while simultaneously attaining a lower solution cost.

To comprehensively investigate the training dynamics of different models, we monitor a suite of key performance metrics during the training process. 
Specifically, for TSP, we track the training loss, training cost, and validation cost to assess the optimization progress. 
For constrained routing problems such as \ac{tspdl}, we additionally monitor the feasible rate on both training and validation sets to evaluate the model's ability to satisfy problem constraints. 
All models are trained for 100 epochs, with each epoch comprising 2500 batches (treating one batch as one training step). 
The resulting learning curves, presented in the following figures, illustrate the evolution of these metrics for the proposed \ac{egam} and baseline algorithms across different routing problems.

As illustrated in \cref{fig:td tsp train} and \cref{fig:td tsp val}, \ac{egam} demonstrates superior training characteristics on the TSP problem compared to \ac{gam} and GATv2. 
The training loss of \ac{egam} exhibits a more rapid convergence toward zero, indicating more effective optimization of the learning objective. 
Concurrently, \ac{egam} achieves a faster reduction in solution cost on both the training and validation sets, suggesting improved learning efficiency. 
These observations collectively demonstrate that \ac{egam} offers advantages in terms of both convergence speed and solution quality.

\begin{figure}[!h]
\centering
\begin{minipage}[b]{.48\textwidth}
  \centering
  \centerline{\includegraphics[width=1.\linewidth]{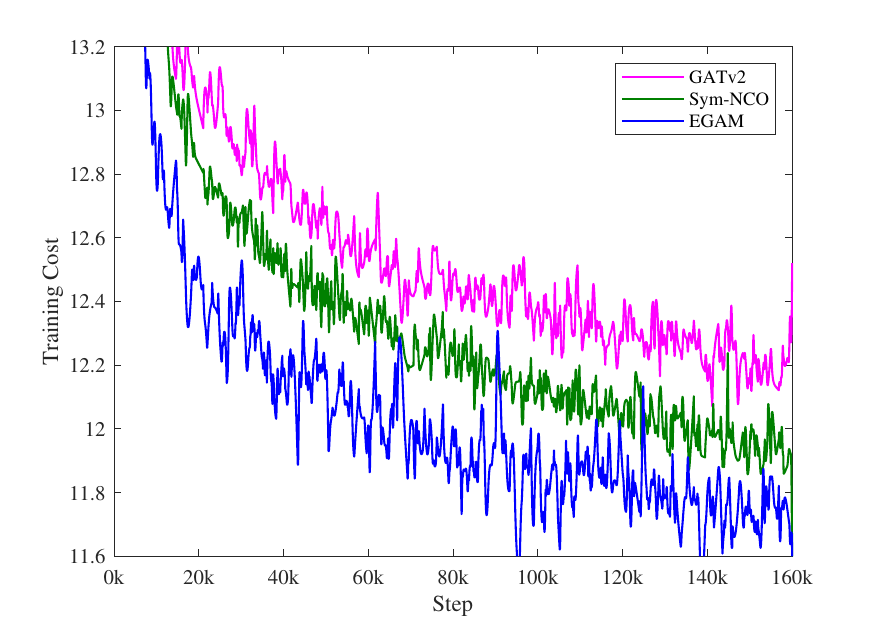}}
  \centerline{(a) Training cost}
  %\medskip
\end{minipage}
\begin{minipage}[b]{.48\textwidth}
  \centering
  \centerline{\includegraphics[width=1.\linewidth]{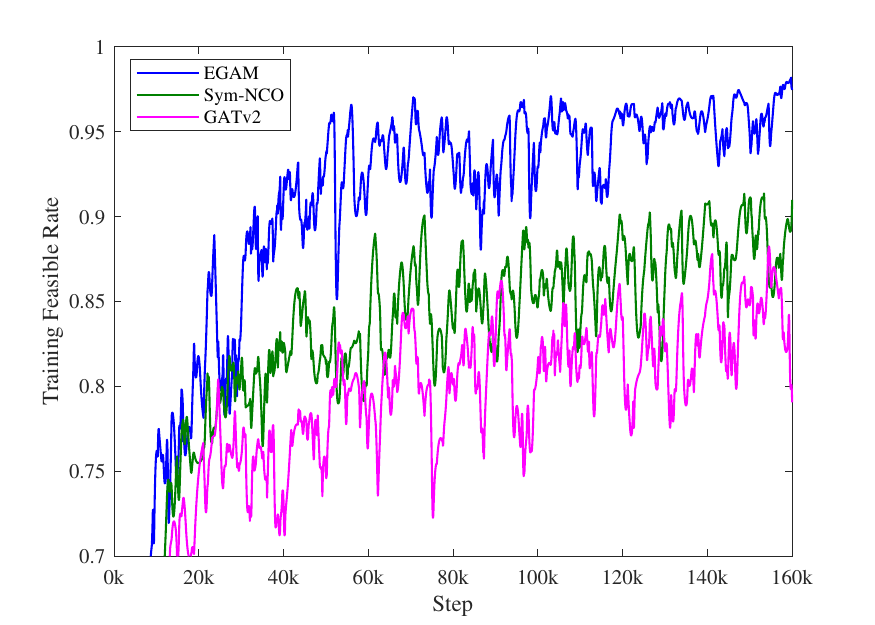}}
  \centerline{(b) Training feasible rate}
  %\medskip
\end{minipage}
% \vspace{-1em}
\caption{Evolution of the training cost and feasible rate for TSPDL}
\label{fig:td tspdl train}
\end{figure}

\begin{figure}[!h]
\centering
\begin{minipage}[b]{.48\textwidth}
  \centering
  \centerline{\includegraphics[width=1.\linewidth]{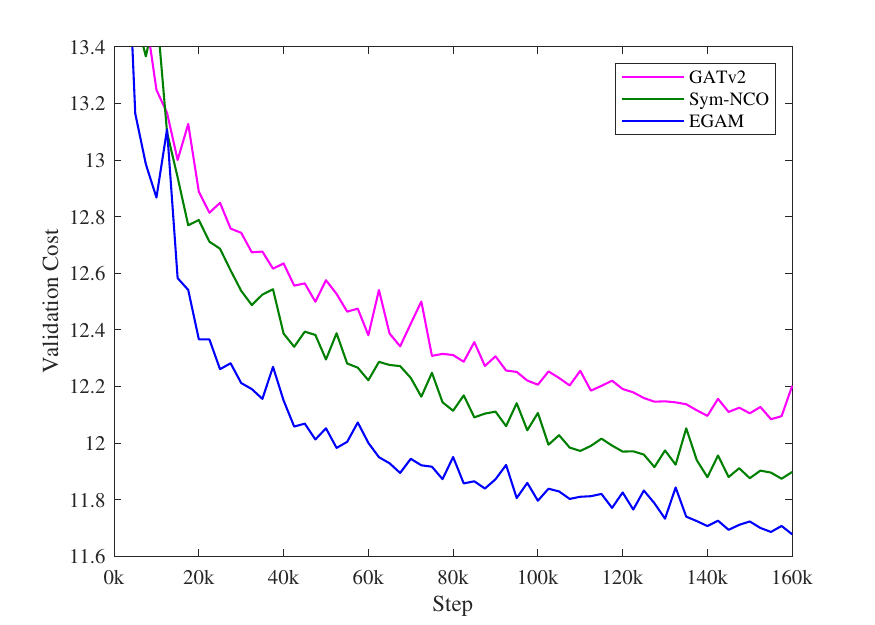}}
  \centerline{(a) Validation cost}
  %\medskip
\end{minipage}
\begin{minipage}[b]{.48\textwidth}
  \centering
  \centerline{\includegraphics[width=1.\linewidth]{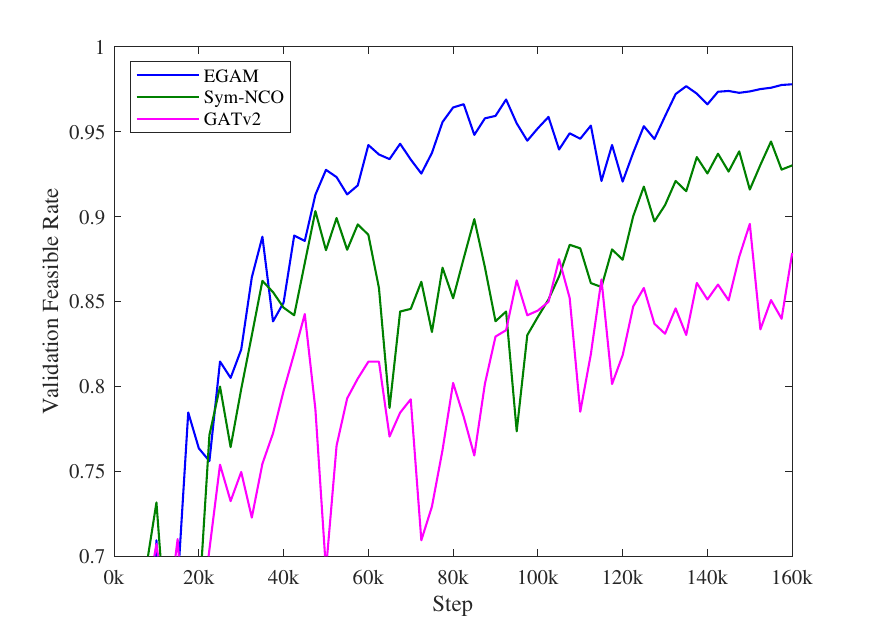}}
  \centerline{(b) Validation feasible rate}
  %\medskip
\end{minipage}
% \vspace{-1em}
\caption{Evolution of the validation cost and feasible rate for TSPDL}
\label{fig:td tspdl val}
\end{figure}

For \ac{tspdl}, as shown in \cref{fig:td tspdl train} and \cref{fig:td tspdl val}, \ac{egam} simultaneously achieves superior performance in both solution quality and constraint satisfaction. 
During training, \ac{egam} achieves a significantly higher feasible rate compared to GATv2 and Sym-NCO, while simultaneously attaining a lower solution cost. 
This indicates that \ac{egam} not only learns to generate feasible solutions more reliably but also optimizes the solution quality within the feasible region. 
The consistent performance gap between \ac{egam} and the baselines on both training and validation sets further validates the model's robustness and generalization capability in handling constrained optimization problems.


\subsection{Visualization of instances}
% tsp-tsptw-vrptw
% To provide an intuitive illustration of routing problems and the results obtained by \ac{nco} solvers, we visualizes the instances and their corresponding solutions in this section. As shown in \cref{fig:instance tsp}, we use dots to represent the nodes to be visited, and arrows to depict the paths. The starting point of the path is marked with a green dot. The objective of the TSP is to find the shortest possible route that visits all nodes, with no constraints imposed on the trajectory. It can be observed that various methods tend to generate a clear cyclic path. Although the cost achieved by \ac{egam} is very close to that of the optimal solution, subtle differences remain between the two paths. This further highlights the high complexity inherent in routing problems. Nonetheless, compared to traditional search-based solvers, \ac{egam} is capable of achieving near-optimal results in significantly less time, demonstrating its practical potential in real-world applications.
To provide an intuitive illustration of routing problems and the solutions obtained by different \ac{nco} solvers, we visualize representative instances and their corresponding solutions in this section. 

\begin{figure}[!h]
\centering
\begin{minipage}[b]{.24\textwidth}
  \centering
  \centerline{\includegraphics[width=.9\linewidth]{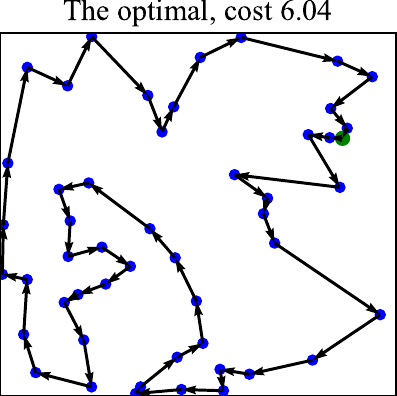}}
  %\medskip
\end{minipage}
\begin{minipage}[b]{.24\textwidth}
  \centering
  \centerline{\includegraphics[width=.9\linewidth]{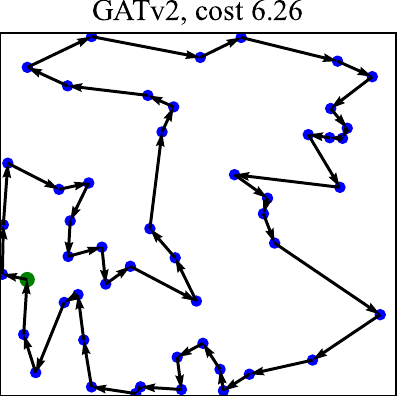}}
  %\medskip
\end{minipage}
\begin{minipage}[b]{.24\textwidth}
  \centering
  \centerline{\includegraphics[width=.9\linewidth]{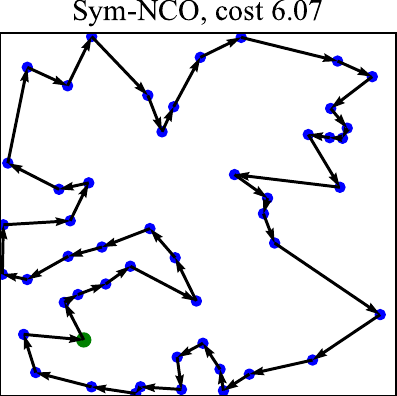}}
  %\medskip
\end{minipage}
\begin{minipage}[b]{.24\textwidth}
  \centering
  \centerline{\includegraphics[width=.9\linewidth]{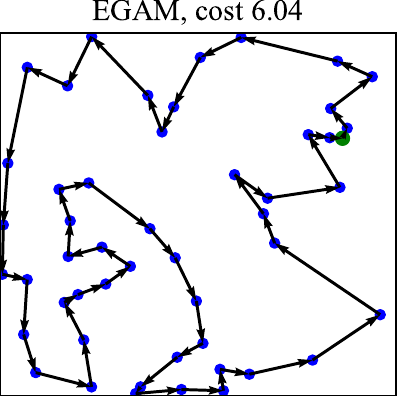}}
  %\medskip
\end{minipage}
% \vspace{-1em}
\caption{Visualization of TSP instance and solutions from different methods}
\label{fig:instance tsp}
\end{figure}

As shown in \cref{fig:instance tsp}, we present the solutions generated by different methods for a representative TSP instance. 
In this visualization, nodes to be visited are represented as dots, and the solution paths are depicted using arrows connecting the nodes. 
The starting point of each path is marked with a green dot to facilitate visual identification. 
The objective of TSP is to find the shortest possible route that visits all nodes exactly once and returns to the starting point, with no additional constraints imposed on the trajectory. 
Visual inspection reveals that all methods generate clear cyclic paths that form closed tours, which is consistent with the problem's structure. 
Notably, while the solution cost achieved by \ac{egam} is very close to that of the optimal solution, subtle differences in the path structure can be observed. 
These differences, though minor in terms of cost, highlight the inherent complexity of routing problems where multiple near-optimal solutions may exist with distinct topological characteristics. 
Compared to traditional search-based solvers that require extensive computational time to find optimal solutions, \ac{egam} demonstrates its capability to achieve near-optimal results efficiently, thereby showcasing its practical potential for real-world applications where computational resources are limited.

\begin{figure}[!h]
\centering
\begin{minipage}[b]{.24\textwidth}
  \centering
  \centerline{\includegraphics[width=.9\linewidth]{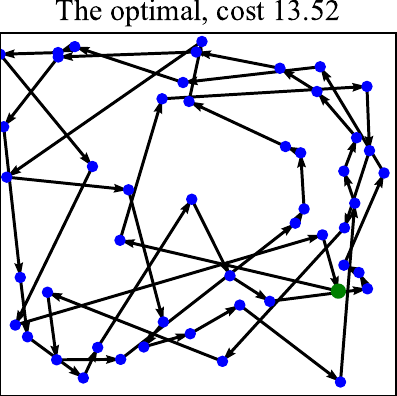}}
  %\medskip
\end{minipage}
\begin{minipage}[b]{.24\textwidth}
  \centering
  \centerline{\includegraphics[width=.9\linewidth]{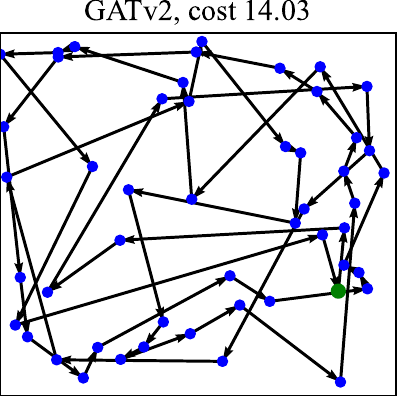}}
  %\medskip
\end{minipage}
\begin{minipage}[b]{.24\textwidth}
  \centering
  \centerline{\includegraphics[width=.9\linewidth]{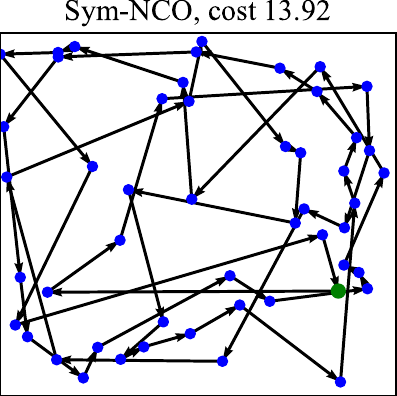}}
  %\medskip
\end{minipage}
\begin{minipage}[b]{.24\textwidth}
  \centering
  \centerline{\includegraphics[width=.9\linewidth]{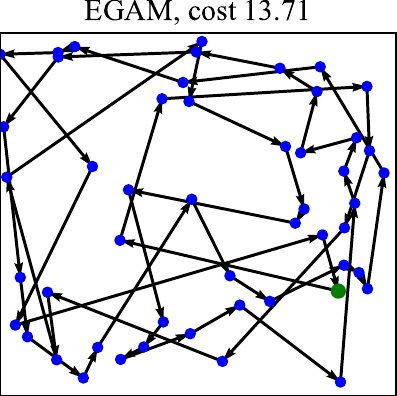}}
  %\medskip
\end{minipage}
% \vspace{-1em}
\caption{Visualization of TSPTW instance and solutions from different methods}
\label{fig:instance tsptw}
\end{figure}

% The visualization of TSPTW results is shown in \cref{fig:instance tsptw}. Compared to TSP, TSPTW adds time window constraints for each node. Only trajectories that satisfy the time windows of all nodes are feasible, and the data generation process ensures the existence of feasible paths. The objective remains to minimize the total travel distance. As illustrated in the figure, both \ac{egam} and the baseline methods (GATv2, Sym-NCO) found solutions within the feasible region, while \ac{egam} achieved the shortest total length.
The visualization of TSPTW results is presented in \cref{fig:instance tsptw}. 
Compared to TSP, TSPTW introduces time window constraints for each node, requiring that the arrival time at each node falls within its specified time window. 
Only trajectories that satisfy the time windows of all nodes are considered feasible solutions. 
The data generation process ensures the existence of at least one feasible path for each instance. 
While the primary objective remains to minimize the total travel distance, the additional constraint significantly complicates the problem by restricting the feasible solution space. 
As illustrated in the figure, all methods, including \ac{egam} and the baseline approaches (GATv2, Sym-NCO), successfully found feasible solutions that satisfy all time window constraints. 
However, \ac{egam} achieved the shortest total travel distance among the feasible solutions, demonstrating its superior ability to optimize within the constrained solution space.

\begin{figure}[!h]
\centering
\begin{minipage}[b]{.24\textwidth}
  \centering
  \centerline{\includegraphics[width=.9\linewidth]{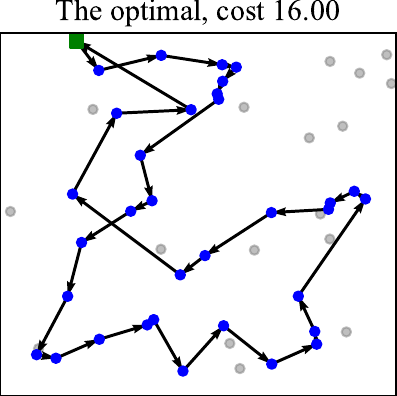}}
  %\medskip
\end{minipage}
\begin{minipage}[b]{.24\textwidth}
  \centering
  \centerline{\includegraphics[width=.9\linewidth]{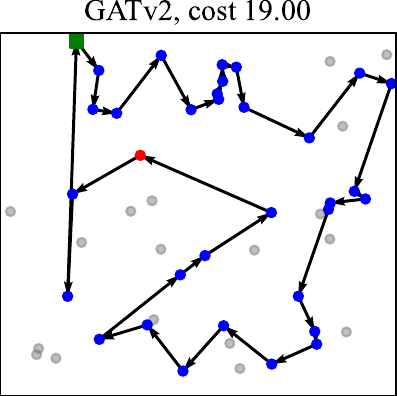}}
  %\medskip
\end{minipage}
\begin{minipage}[b]{.24\textwidth}
  \centering
  \centerline{\includegraphics[width=.9\linewidth]{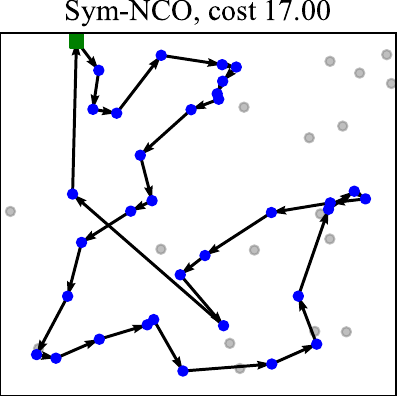}}
  %\medskip
\end{minipage}
\begin{minipage}[b]{.24\textwidth}
  \centering
  \centerline{\includegraphics[width=.9\linewidth]{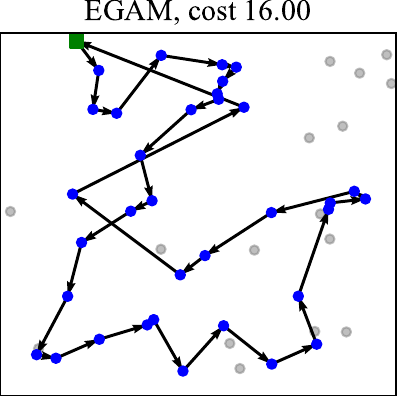}}
  %\medskip
\end{minipage}
% \vspace{-1em}
\caption{Visualization of VRPTW instance and solutions from different methods}
\label{fig:instance vrptw}
\end{figure}

% For VRPTW, we consider a new problem setting as described in \cref{appendixsubsec:vrptw}. In this modified VRPTW, the objective shifts to visiting as many nodes as possible within their respective time windows. As shown in \cref{fig:instance vrptw}, blue points indicate successfully visited nodes, red points represent failed visits, and gray points denote unvisited nodes. The problem's cost is defined as the number of unsuccessfully visited nodes.
% For the instance shown in the figure, \ac{egam}, Sym-NCO, and GATv2 successfully visited 34, 33, and 31 nodes, respectively.
For VRPTW, we consider a novel problem setting as described in \cref{appendixsubsec:vrptw}. 
In this modified VRPTW formulation, the objective shifts from minimizing travel distance to maximizing the number of nodes successfully visited within their respective time windows. 
This problem setting is particularly relevant for scenarios where service coverage is prioritized over travel efficiency. As shown in \cref{fig:instance vrptw}, the visualization employs a color-coded scheme to distinguish different node states: blue points indicate successfully visited nodes that satisfy their time window constraints, red points represent nodes where visits were attempted but failed due to constraint violations, and gray points denote nodes that were not visited at all. 
The problem's cost is defined as the number of unsuccessfully visited nodes (i.e., the sum of failed and unvisited nodes), making the objective equivalent to minimizing this cost or equivalently maximizing the number of successfully visited nodes. 
For the representative instance shown in the figure, \ac{egam} successfully visited 34 nodes, outperforming Sym-NCO (33 nodes) and GATv2 (31 nodes).

\subsection{Impact of the number of parameters}
% vrptw sym-3/6/12 egam-3/4
% To analyze the efficiency of the proposed model in utilizing learnable parameters, we conducted experiments to investigate the impact of varying numbers of layers on model performance.
% tabel 层数-参数量-性能
To assess the parameter efficiency of the proposed model, we conducted experiments to examine the impact of varying the number of encoder layers on model performance. 
We compare different layer configurations of \ac{egam} and Sym-NCO on the VRPTW, analyzing both the number of parameters and the resulting performance, as summarized in \cref{table:parameters}. 
\begin{table}[h!]
\caption{Impact of the number of parameters on Sym-NCO and EGAM}
\label{table:parameters}
\vskip 0.15in
\begin{center}
\begin{small}
\begin{tabular}{l|ccc}
\toprule
{Method} & Layers & Parameters & Performance on VRPTW\\
\midrule
Sym-NCO  &3  &$0.69\times 10^6$  &  20.38 \\
Sym-NCO  &6 &$1.29\times 10^6$  &  19.97 \\
Sym-NCO  &9  &$1.88\times 10^6$  &  19.89\\
Sym-NCO  &12  &$2.47\times 10^6$  &   -\\
\midrule
EGAM &3 &$1.70\times 10^6$ & 19.82\\
EGAM &4 &$2.16\times 10^6$ & 19.60\\
\bottomrule
\end{tabular}
% \end{sc}
\end{small}
\end{center}
\vskip -0.1in
\end{table}

% 图片 层数-算法曲线
\begin{figure}[!h]
\begin{minipage}[b]{1.\textwidth}
  \centering
  \centerline{\includegraphics[width=.7\linewidth]{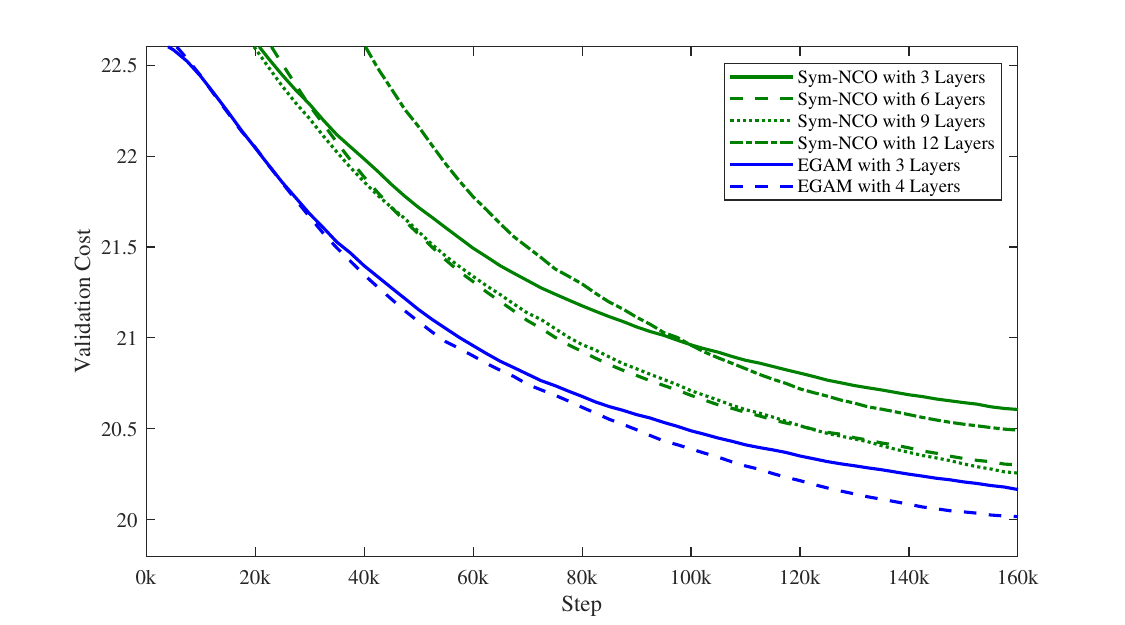}}
%   \centerline{(c) Validation cost}
  %\medskip
\end{minipage}
\vspace{-1em}
\caption{Validation cost curves during training for models with varying numbers of layers. For better clarity, an exponential smoothing with a coefficient of 0.9 was applied.}
\label{fig:params}
\end{figure}

As shown in \cref{table:parameters}, \ac{egam} requires more parameters per layer due to the simultaneous update of node and edge embeddings in each encoder layer. 
With a similar number of parameters, the 3-layer \ac{egam} outperforms the 9-layer Sym-NCO. 
Furthermore, when the number of parameters is large, the performance improvement from 9 layers to 6 layers in Sym-NCO is minimal. 
In contrast, the 4-layer \ac{egam} shows a clear improvement over the 3-layer version. We attribute this to the network structure (as shown in \cref{fig:model}), where \ac{egam} has a shallower depth for parameter gradient backpropagation, making it easier to train.
We also present the validation cost curves during training for models with varying numbers of layers in \cref{fig:params}. 
From the figure, it is evident that the 12-layer Sym-NCO encounters significant training difficulties and struggles to converge (even after adjusting the learning rate, the issue persists). 
Additionally, the training cost for \ac{egam} decreases more rapidly in the early stages compared to Sym-NCO. 
The two curves for \ac{egam} consistently lie below the curves for Sym-NCO.
These results indicate that our model architecture is more conducive to efficient parameter training, achieving faster convergence and better performance even with the same number of parameters.

\subsection{Generalization performance}
\begin{figure}[!h]
\centering
\begin{minipage}[b]{.45\textwidth}
  \centering
  \centerline{\includegraphics[width=.9\linewidth]{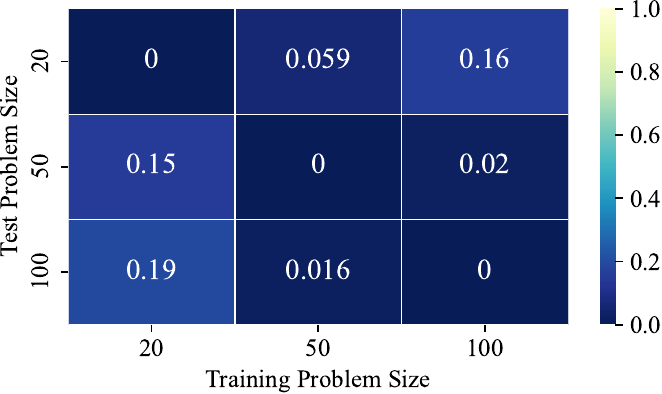}}
  \centerline{(a)}
  %\medskip
\end{minipage}
\begin{minipage}[b]{.45\textwidth}
  \centering
  \centerline{\includegraphics[width=.9\linewidth]{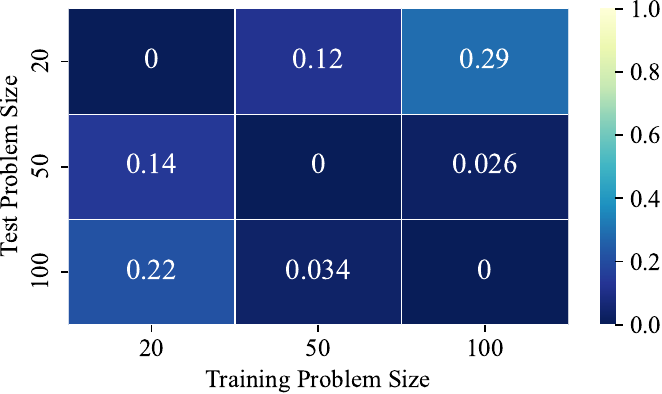}}
  \centerline{(b)}
  %\medskip
\end{minipage}
% \vspace{-1em}
\caption{Generalization tests of EGAM (a) and Sym-NCO (b) trained and tested on VRPTW
with various scales}
\label{fig:generalization test}
\end{figure}

% TSPTW/VRPTW 20/50/100
% We evaluated the generalization capability of the proposed \ac{egam} across different problem scales by generalization tests. Specifically, we considered three scales—20, 50, and 100—and evaluated models obtained from corresponding datasets on other scales.
% Table \cref{table:general} provides the test results for models of the same problem scale, with the diagonal entries of \cref{fig:generalization test} representing the average cost for each model.
% Table \cref{table:general} provides the test results for models of the same problem scale, with the diagonal entries of \cref{fig:generalization test} corresponding to the average cost for each model.
We evaluated the generalization capability of the proposed \ac{egam} across different problem scales using generalization tests. 
Specifically, we considered three scales—20, 50, and 100—and assessed the models trained on corresponding datasets by testing them on datasets of other scales.
\cref{fig:generalization test} presents a heatmap showing the performance gap of models trained on datasets of different scales but evaluated on the same problem scale. 
\cref{table:general} provides the test results for models of the same problem scale. The average cost for each model, corresponding to the diagonal entries of \cref{fig:generalization test}, is listed in the table.

\begin{table}[h!]
\caption{Performance of EGAM and Sym-NCO on VRPTW with different scales}
\label{table:general}
\vskip 0.15in
\begin{center}
\begin{small}
\begin{tabular}{l|ccc}
\toprule
{Problem size} & 20 & 50 & 100\\
\midrule
Sym-NCO &4.65 &19.97 &52.92 \\
EGAM &4.58 &19.60 &52.53 \\
\bottomrule
\end{tabular}
% \end{sc}
\end{small}
\end{center}
\vskip -0.1in
\end{table}

As shown in \cref{fig:generalization test} (a) and (b), the diagonal elements are 0 because the models trained and tested on the same problem scale always perform the best. 
Comparing the two figures, it is evident that the performance gap between different \ac{egam} models is smaller, indicating stronger generalization capability. 
The data in \cref{table:general} further confirms that our model consistently outperforms the baseline across different problem scales.

\section{Future Work}
\label{appendixsec:future}

In this paper, we introduced the foundational model of \ac{egam} within an autoregressive framework. Notably, \ac{egam} demonstrates strong scalability. In future work, we aim to extend the application of \ac{egam} to a broader range of scenarios, specifically in the following directions:
\begin{itemize}
    \item Integrating improved methods tailored for NCO to enhance performance on specific problem types and larger-scale instances.
    \item Extending its application to a wider variety of routing problems, such as edge-based route planning.
    \item Exploring non-autoregressive architectures. By modifying the decoder, a non-autoregressive version of \ac{egam} will output routing planning heatmaps, which can be trained using supervised learning.
    \item Generalizing the approach beyond routing problems to tackle a wider variety of combinatorial optimization tasks. Considering other combinatorial optimization problems, such as the Maximal Independent Set problem, many of which can also be represented as graph-based problems, we plan to leverage \ac{egam} to efficiently address these challenges and establish a universal framework for graph-structured problems.
\end{itemize}

\section{Licenses and Assets}
\label{appendixsec:licenses}

In this work, we adopt existing routing solvers and neural combinatorial optimization methods as benchmarks for our experiments. 
The implementation of our innovations references the code frameworks developed by other researchers. 
The assets used in this work, along with their corresponding licenses, are listed in Table~\ref{table:assets}, all of which are available for academic use.
Furthermore, we are also happy to release the source code developed in this study under the MIT License in the future, to facilitate communication and collaboration with other researchers.

\begin{table}[h!]
\caption{Assets used in this work}
\label{table:assets}
\vskip 0.15in
\begin{center}
\begin{small}
\begin{tabular}{l|l|l}
\toprule
Asset & License & URL \\
\midrule
Gurobi  &Academic license for non-commercial use & \url{https://www.gurobi.com/solutions/gurobi-optimizer/}   \\
OR-Tools  & Apache-2.0 license  & \url{https://developers.google.com/optimization/routing/}\\
LKH-3  & Available for academic use  & \url{http://webhotel4.ruc.dk/~keld/research/LKH-3/}   \\
% GATv2 & Available for academic use & \url{https://github.com/tech-srl/how_attentive_are_gats}\\
% Sym-NCO & Available for academic use& \url{https://github.com/alstn12088/Sym-NCO}\\
GAM & MIT license & \url{https://github.com/wouterkool/attention-learn-to-route}\\
POMO & MIT license & \url{https://github.com/yd-kwon/POMO}\\
PIP & MIT license & \url{https://github.com/jieyibi/PIP-constraint}\\
\bottomrule
\end{tabular}
% \end{sc}
\end{small}
\end{center}
\vskip -0.1in
\end{table}
%%%%%%%%%%%%%%%%%%%%%%%%%%%%%%%%%%%%%%%%%%%%%%%%%%%%%%%%%%%%%%%%%%%%%%%%%%%%%%%
%%%%%%%%%%%%%%%%%%%%%%%%%%%%%%%%%%%%%%%%%%%%%%%%%%%%%%%%%%%%%%%%%%%%%%%%%%%%%%%

\end{document}

%% file: Wgroup-Teaching-Preamble.tex
\usepackage{bm}
\usepackage{upgreek}

\DeclareMathAlphabet{\mathsfbr}{OT1}{cmss}{m}{n}%for math sans serif (cmss)
\SetMathAlphabet{\mathsfbr}{bold}{OT1}{cmss}{bx}{n}%for math sans serif (cmss)
\DeclareRobustCommand{\msf}[1]{%
  \ifcat\noexpand#1\relax\msfgreek{#1}\else\mathsfbr{#1}\fi%for math sans serif (cmss)
}
\makeatletter
\newcommand{\msfgreek}[1]{\csname s\expandafter\@gobble\string#1\endcsname}
\makeatother

\DeclareFontEncoding{LGR}{}{} % or load \usepackage{textgreek}
\DeclareSymbolFont{sfgreek}{LGR}{cmss}{m}{n}
\SetSymbolFont{sfgreek}{bold}{LGR}{cmss}{bx}{n}
\DeclareMathSymbol{\salpha}{\mathord}{sfgreek}{`a}
\DeclareMathSymbol{\sbeta}{\mathord}{sfgreek}{`b}
\DeclareMathSymbol{\sgamma}{\mathord}{sfgreek}{`g}
\DeclareMathSymbol{\sdelta}{\mathord}{sfgreek}{`d}
\DeclareMathSymbol{\sepsilon}{\mathord}{sfgreek}{`e}
\DeclareMathSymbol{\szeta}{\mathord}{sfgreek}{`z}
\DeclareMathSymbol{\seta}{\mathord}{sfgreek}{`h}
\DeclareMathSymbol{\stheta}{\mathord}{sfgreek}{`j}
\DeclareMathSymbol{\siota}{\mathord}{sfgreek}{`i}
\DeclareMathSymbol{\skappa}{\mathord}{sfgreek}{`k}
\DeclareMathSymbol{\slambda}{\mathord}{sfgreek}{`l}
\DeclareMathSymbol{\smu}{\mathord}{sfgreek}{`m}
\DeclareMathSymbol{\snu}{\mathord}{sfgreek}{`n}
\DeclareMathSymbol{\sxi}{\mathord}{sfgreek}{`x}
\DeclareMathSymbol{\somicron}{\mathord}{sfgreek}{`o}
\DeclareMathSymbol{\spi}{\mathord}{sfgreek}{`p}
\DeclareMathSymbol{\srho}{\mathord}{sfgreek}{`r}
\DeclareMathSymbol{\ssigma}{\mathord}{sfgreek}{`s}
\DeclareMathSymbol{\stau}{\mathord}{sfgreek}{`t}
\DeclareMathSymbol{\supsilon}{\mathord}{sfgreek}{`u}
\DeclareMathSymbol{\sphi}{\mathord}{sfgreek}{`f}
\DeclareMathSymbol{\schi}{\mathord}{sfgreek}{`q}
\DeclareMathSymbol{\spsi}{\mathord}{sfgreek}{`y}
\DeclareMathSymbol{\somega}{\mathord}{sfgreek}{`w}

\DeclareMathSymbol{\svarsigma}{\mathord}{sfgreek}{`c}

\DeclareMathSymbol{\sGamma}{\mathalpha}{sfgreek}{`G}
\DeclareMathSymbol{\sDelta}{\mathalpha}{sfgreek}{`D}
\DeclareMathSymbol{\sTheta}{\mathalpha}{sfgreek}{`J}
\DeclareMathSymbol{\sLambda}{\mathalpha}{sfgreek}{`L}
\DeclareMathSymbol{\sXi}{\mathalpha}{sfgreek}{`X}
\DeclareMathSymbol{\sPi}{\mathalpha}{sfgreek}{`P}
\DeclareMathSymbol{\sSigma}{\mathalpha}{sfgreek}{`S}
\DeclareMathSymbol{\sUpsilon}{\mathalpha}{sfgreek}{`U}
\DeclareMathSymbol{\sPhi}{\mathalpha}{sfgreek}{`F}
\DeclareMathSymbol{\sPsi}{\mathalpha}{sfgreek}{`Y}
\DeclareMathSymbol{\sOmega}{\mathalpha}{sfgreek}{`W}

\DeclareRobustCommand{\mcal}[1]{%
  \ifcat\noexpand#1\relax\mathnormal{#1}\else\cal{#1}\fi
}
\DeclareRobustCommand{\BM}[1]{%
  \ifcat\noexpand#1\relax\bm{\boldUppercaseItalicGreek{#1}}\else\bm{#1}\fi
}
\makeatletter
\newcommand{\boldUppercaseItalicGreek}[1]{\csname var\expandafter\@gobble\string#1\endcsname}
\makeatother
\newcommand{\RV}[1]{\bm{\MakeLowercase{\msf{#1}}}}
\newcommand{\RM}[1]{\bm{\MakeUppercase{\msf{#1}}}}

\newcommand{\V}[1]{\bm{#1}}
\newcommand{\M}[1]{\BM{#1}}
\newcommand{\Set}[1]{\mcal{#1}}